\pdfoutput=1
\documentclass[suppldata]{interact}
\usepackage[caption=false]{subfig}
\usepackage{amsmath}
\usepackage[nopatch]{microtype}
\usepackage{booktabs}
\usepackage{setspace}
\usepackage[section]{placeins}
\usepackage{float}
\usepackage[utf8]{inputenc}
\usepackage{graphicx}
\usepackage{longtable}
\usepackage[most]{tcolorbox}
\usepackage{pdflscape}
\usepackage{rotating}
\usepackage{geometry}
\newtcolorbox{promptbox}[2]{breakable,colback=#1!5!white,colframe=#1!60!black,title=#2,fonttitle=\bfseries,left=1.5mm,right=1.5mm,top=1mm,bottom=1mm,boxrule=0.5pt}

\begin{document}
\doublespacing

\title{
Navigating the Prompt Space: Improving LLM Classification of Social Science Texts Through Prompt Engineering}

\author{
\name{Erkan Gunes\textsuperscript{1}, Christoffer Florczak\textsuperscript{2}, and Tevfik Murat Yildirim\textsuperscript{3}}
\affil{\textsuperscript{1}Constructor University}
\affil{\textsuperscript{2}Aalborg University}
\affil{\textsuperscript{3}University of Stavanger}
}

\maketitle

\begin{abstract}
Recent developments in text classification using Large Language Models (LLMs) in the social sciences suggest that costs can be cut significantly, while performance can sometimes rival existing computational methods. However, with a wide variance in performance in current tests, we move to the question of how to maximize performance. In this paper, we focus on prompt context as a possible avenue for increasing accuracy by systematically varying three aspects of prompt engineering: label descriptions, instructional nudges, and few shot examples. Across two different examples, our tests illustrate that a minimal increase in prompt context yields the highest increase in performance, while further increases in context only tend to yield marginal performance increases thereafter. Alarmingly, increasing prompt context sometimes decreases accuracy. Furthermore, our tests suggest substantial heterogeneity across models, tasks, and batch size, underlining the need for individual validation of each LLM coding task rather than reliance on general rules.
\end{abstract}
\section{Introduction}

Text annotation using instruction-tuned large language models (LLMs) has emerged as an alternative to training custom supervised models for text annotation in social science research. While findings on accuracy were initially mixed (Ziems et al., 2024; Kristensen-McLachlan et. al., 2025), there is now growing evidence that for some tasks, they could offer satisfactory accuracy when used out-of-the-box with minimal context provided by the user, and therefore have significant potential to replace existing computational tools when fine-tuned for a specific task or with more sophisticated prompting or context-building strategies (Liu and Shi, 2024; Bosley, 2025, Halterman and Keith, 2025). As base accuracy has been established against benchmarks, the cost cutting potential is clear, and given that models are likely to improve in the future, the questions around LLMs for text annotation should change nature. Rather than exploring if they are valuable text annotators, the social science community could benefit by exploring how to optimize LLMs for text annotation. Focusing on one avenue, namely prompt engineering, we seek to answer this question by providing an initial test of how researcher input might affect LLM text classification performance.

Prompt engineering is an obvious first avenue to test optimization, as there is large control for the individual researcher, ample opportunity to experiment as the costs are usually low, and one cannot avoid prompting the model, thus making it a necessary step for all researchers. This is, however, complicated by the fact that there are almost an infinite number of possibilities for prompt configurations that can be submitted to a model. To narrow this down, we focus on three areas of prompting. First, we analyze the depth of candidate label description, where we vary if the labels are only given to the model, or if the labels are also explained to the model. Second, we experiment with instructional nudges, where we vary if small notes clarifying some nuances about the classification task and setting expectations about LLM output are included in the prompt. Third, we vary whether the model is presented with few-shot examples, or whether the model is given no examples. Additionally, we also vary the size of the input text batch and the model used to further evaluate how prompt engineering interacts with other specifications to affect LLMs’ label prediction. We thus provide a fairly comprehensive set of tests on common prompt engineering dilemmas faced by social scientists.

We focus on two classification tasks relevant to social science research and present them as two separate studies in this work. The first study draws on a political science specific task and involves classification of congressional bill titles into 21 policy issue topic categories. The second study has relevance to many fields such as psychology and sociology, as well as political science, and involves the classification of open-ended survey responses into emotional vs. neutral categories. Results from our experiments spanning all combinations of the above-mentioned context components suggest that researchers can achieve substantial accuracy gains when they go beyond the most minimalist configuration, but there is not a uniform pattern of improvement, e.g. increasing relevant contextual information within the prompt does not linearly increase accuracy. We also find that presenting a single input text within a prompt is usually the suboptimal strategy in terms of performance as well as the more costly option, thus suggesting that batch classification will be the superior choice in many cases. Additionally, our results from the emotionality classification provide evidence that deterministic LLM outputs are not guaranteed even when the temperature parameter is set to zero, and inter-request variation with the exact same prompt content depends on model and prompt attributes. Taken together, our results highlight how performance varies across prompt configuration, task, and model, and there is therefore a need for individual case specific validation tests when performing text annotation using LLMs if researchers want to harness the efficiency of LLM based text classification.

This article proceeds as follows. In the second section, we review recent research on text annotation with LLMs and discuss the shifting focus from model benchmarking to navigating the prompt space for understanding and improving model behavior. In the third section, we introduce the experimental setup of our two studies, and present the results from our prompt engineering experiments. In the fourth section, we discuss what we can learn from our results about understanding and improving LLMs text annotation performance. In the fifth section, we conclude with some general lessons we can draw from our two studies and discuss future research.

\section{Related Works}

Scholarly interest in the use of instruction-tuned LLMs for various scientific research tasks has increased sharply since the release of ChatGPT by OpenAI. In social science research, this has been especially evident in computational text annotation literature. Until recently, this branch of computational social science was dominated by efforts to develop custom supervised models for specific tasks. The flexibility, the ease-of-use, and the promising accuracy performance of the latest instruction-tuned LLMs has shifted the field’s attention away from training and fine-tuning models toward text annotation using in-context learning methods enabled by instruction-tuned LLMs.

This shift, in turn, motivated an initial wave of benchmarking studies that compared zero-shot LLM-based annotation to established baselines, including supervised models and human annotators. The resulting evidence from this initial wave of studies helped clarify if LLMs could replace human coders and the more conventional computational tools and methods, or if they could be a collaborator to humans. Gilardi, Alizadeh, and Kubli (2023) showed GPT 3.5 outperforms crowd workers on several text classification tasks such as stance detection and frame detection, and does it much more cheaply. Similarly, Törnberg (2024) experimented with GPT 4 on a political affiliation detection task and found that it overperforms both supervised models and human coders. That work also highlights groundbreaking implications of LLMs for the future of computational text annotation in social science research. Experimenting with a larger number of models on a more complicated multiclass text classification task, which involves classification of congressional bill titles and descriptions of congressional hearings into policy issue categories, Gunes and Florczak (2025) reported LLMs can sometimes rival human coders and supervised models. They also found that best performance occurs when humans and LLMs collaborate in a way where LLMs label cases only when multiple LLMs agree, and humans label cases when the LLMs disagree.

While this first wave of studies hinted that annotation performance could be sensitive to prompt design choices, most research did not systematically explore the prompt space to identify optimal prompt configurations or to develop procedures for analyzing and reporting prompt-related sensitivities. With growing awareness that annotation quality depends on prompt design choices (Atreja et al. , 2025), especially the size and structure of contextual information included in the prompt and the reasoning strategies the model is instructed to use, recent research on text annotation with LLMs has started focusing more on navigating the prompt space than on comparing the capabilities of a set of models to humans and conventional computational tools.

LLMs learn from contextual information presented in the prompt (Brown et al., 2020). A central problem in research on prompt engineering for text annotation is how to improve the quality of LLM output by providing the right contextual information. One contextual component researchers often manipulate in prompt engineering experiments is few-shot examples. Providing a few examples for each candidate label in a classification task can often help boost accuracy. Bosley (2025) demonstrates that, as the number of in-context examples increases, LLMs can approach expert-level fidelity on an intricate multidimensional coding scheme, the Discourse Quality Index for deliberative debates. Performance improved markedly when moving from zero-shot to a handful of examples, and then plateaued around 25-50 examples, suggesting diminishing returns beyond a certain context length. Liu and Shi (2024) also emphasize more systematic prompt optimization pipelines that go beyond manually selecting exemplars. They introduce PoliPrompt, a three-stage framework that iteratively refines prompts, dynamically selects relevant in-context examples, and uses a consensus mechanism to reconcile outputs across prompts or models. They report sizable gains in classification performance alongside large reductions in annotation costs.

Getting the number and the kind of few-shot examples right is a central objective in more generic prompt optimization frameworks like DSPy, as well (Khattab et al, 2023). In DSPy, users tell the model what the inputs and outputs are for their task, choose additional parameters such as an optimizer ,e.g. MIPRO (Opsahl-Ong, 2024) or GEPA (Agrawal, 2025), and an evaluation metric, and then run DSPy to search for a high performing prompt configuration under that metric. In a text classification task, the input is a short natural language text or document, and the output is a label describing the category the text belongs to. DSPy requires a dataset with ground truth labels, which it uses to compare different possible prompt configurations and find the best performing one on the user-defined input-output formulation. While DSPy offers a wide range of controllable modules, including mechanisms for selecting or synthesizing few-shot examples, its relatively steep learning curve and higher technical barriers to entry may make it impractical for many social scientists. In such cases, researchers may seek greater manual control over prompt content and prioritize domain expertise rather than relying on more automated prompt optimization to determine what information is presented to the model.

Beyond few-shot prompting, researchers are developing systematic methods to discover optimal prompting strategies. Reich et al. (2025) propose the HALC pipeline as a general and codebook-driven approach for automated coding in social science research. HALC allows integrating various prompting strategies (e.g. role prompting, chain-of-thought prompting, context information) and evaluates them empirically at scale. They provide results from experiments where they tried over 1500 candidate prompts in more than two million requests to a local LLM, where the results highlight the importance of high-quality ground-truth data, the choice of prompting strategy, and using a majority decision procedure with multiple LLMs.

While the focus still is largely on highly automated frameworks and pipelines for text annotation, Than et al. (2025) explore using generative LLMs in a more interactive manner. They argue that LLMs, prompted in natural language, can mimic the inductive and iterative process of human qualitative analysis to some extent. Their experiments with multiple models and prompt styles showed that LLMs can replicate many of the outcomes of traditional hand-coding, coming close to the accuracy of human-coded themes and even matching prior machine learning baselines. More intriguingly, they highlight how a researcher can engage in a back-and-forth dialogue with an LLM, such as asking the model to explain why a text was labeled a certain way, or to suggest new labels. This points to an emerging paradigm of AI-in-the-loop qualitative research, where the LLM is not just an automated coder but also a conversational partner that can help analysts explore data.

The emerging prompt engineering in computational text annotation literature has already identified various effective prompt components and prompt building strategies. In this study, we contribute to the exploration of the vast prompt space by shifting attention to some structural prompt design elements that can help improve accuracy performance of LLMs and to some sensitivity issues that have not been paid enough systematical attention. In the next section, we introduce these elements and the experimental setup in greater detail.

\section{Experiments on Context-Building for Text Classification Tasks}

In this work, we deal with two different classification problems, which we present as two separate studies. The first study is a multiclass classification problem where congressional bill titles are classified into policy issue topics. We use the Congressional Bills dataset (Wilkerson et al., 2025) from the Comparative Agendas Project (CAP). The dataset contains human coded congressional bill titles. The CAP topic scheme offers a comprehensive set of policy issue topic classes, where there are 21 unique classes. Classification of policy documents into the CAP topics is a challenging multiclass classification problem given the large number of classes and vague semantic boundaries between some classes. The second study focuses on a binary classification problem where we work on emotionality detection in open-ended survey responses from American National Election Studies surveys (Yildirim, Chang and Williams 2026). A statement belongs to the “emotional” class if it contains any emotional expression, or it belongs to the “neutral” class if it features no emotional expression.

Our primary goal in this work is not comparing classification capabilities of specific LLMs, but to explore a wide range of prompt design configurations and consequences on classification accuracy and reproducibility. However, to avoid overgeneralization about LLM behavior under the configurations we test and to find out if cross-configuration patterns we observe are model dependent, we implemented our experiments with more than one LLM. In both studies, we use two different instruction-tuned LLMs. We use GPT 4o from OpenAI and Gemini 2.0 Flash from Google, which are relatively fast and cheap models with high scores across many performance benchmarks. 

In the following subsections, we present details about datasets, prompt configuration components and the results from experiments for both studies.

\subsection{Study 1: Classifying Congressional Bill Titles into Policy Issue Topics}

A large body of work within the literature on policy agendas focuses on attention allocation dynamics in policy making processes (Jones and Baumgartner 2005; Baumgartner and Jones 2015). Identifying attention allocation distribution across issue categories at a given time and policy venue is essential for describing the dynamics of attention allocation in policy making processes and investigating the implications for policy outputs (Baumgartner et al. 2009). Researchers have long been processing various policy documents and manually classifying them into a policy issue category. CAP offers the most widely used topic scheme for policy issue classification with more than 21 major issue topics and numerous issue subtopics under each major topic (Baumgartner, Jones and Mortensen 2018). There has been a major effort to automate issue topic classification as large advances in natural language processing have revolutionized how researchers can process unstructured text data (Sebők and Kacsuk, 2021; Baumgartner, Bevan, and Sebők, 2026). More recently, there have been efforts to automate issue topic classification using instruction-tuned LLMs (Gunes and Florczak, 2025), where the results showed promising accuracy performance rivaling custom-trained supervised learning models. In this first study, we explore further how we can get better classification performance from instruction-tuned LLMs when they are used for a highly complex classification problem as the policy issue classification.

In our policy issue classification experiments, we use the prompt structure depicted in Figure 1. The three small boxes in the center represent the contextual information components we systematically vary in the experiments. The left box is the “label description” component. When this component is present in a prompt configuration, we present candidate labels along with a description of the kinds of policy issues that belong to that CAP major topic category. When this component is absent in a prompt configuration, we only include the list of candidate labels without any description. Label descriptions come from the master codebook of the CAP. In the congressional bills dataset, there is an additional class, named “private bills” which does not exist in the default CAP scheme. Private bills refer to proposals that are proposed to solve a problem concerning a private individual, typically addressing a unique situation not covered by public laws. The label description for that class comes from the official website of the U.S. Senate.\footnote{https://www.senate.gov/legislative/common/briefing/leg\_laws\_acts.htm} The box in the middle of the center part is the “instructional nudges” component. These are notes that serve as instructional nudges, and they are informed by domain expertise and small-scale trials with models. For example, abortion related bill titles belong to the civil rights category in the CAP codebook, but the models we used were often assigning them the “health” label. We included a note to explain how abortion related titles must be handled by the model. Such notes focus on instructions that are very specific in nature, and they provide extra context which cannot be presented in the “label descriptions” or “few-shot examples” parts. When this component is present in a prompt configuration, we show such instructions in the middle of the prompt, and we show no such instructions when it is absent. The right box in the center is the “few-shot examples” component, which shows the LLM model a few congressional bill title examples for each unique class.

\begin{figure}[htbp]
\centering
\includegraphics[width=0.95\linewidth]{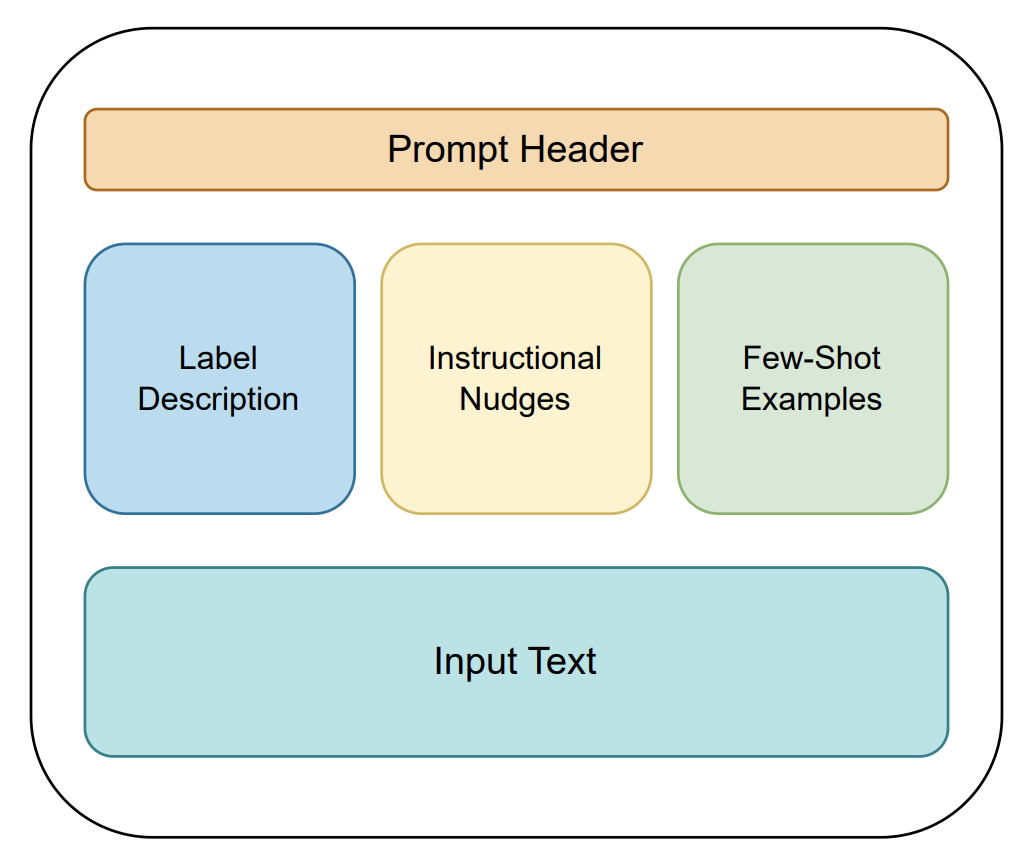}
\caption{Prompt components that vary across experimental configurations}
\end{figure}

The examples come from the part of the full Congressional Bills dataset which did not end up in the random sample we used in the experiments. When this component is absent in a prompt configuration, we do not show any example titles for any major topic category. The bottom component in Figure 1 is the input text component, which is congressional bill titles in our case. Batch size is the dimension that varies in this component of the prompt. We use batch sizes of 1, 10, 100, 500, and 1000. When the batch size is 1, we show a single bill title to be classified. When it is greater than 1, we show multiple bill titles in a single prompt and ask the LLM model to return a list of labels generated in a format specified within the prompt. We show full prompt examples, where all the context components are present and have a batch size of 1, in Appendix 1.

We present the results from our prompt engineering experiments with the congressional bills data in Figure 2 and Figure 3. They show weighted F1 scores for all possible configurations of the contextual information components described above, five different input text batch sizes, and two different LLM models. Each vertical section separated by dashed lines correspond to a different combination of the three components, which are respectively “label descriptions”, “instructional nudges”, and “few-shot examples”. The plus and minus signs in the x-axis labels show whether a component is present or absent in a configuration. For example, in the configuration denoted by (+,-,+), “instructional nudges” component does not exist, but the other two components are present.

\begin{figure}[htbp]
\centering
\includegraphics[width=0.95\linewidth]{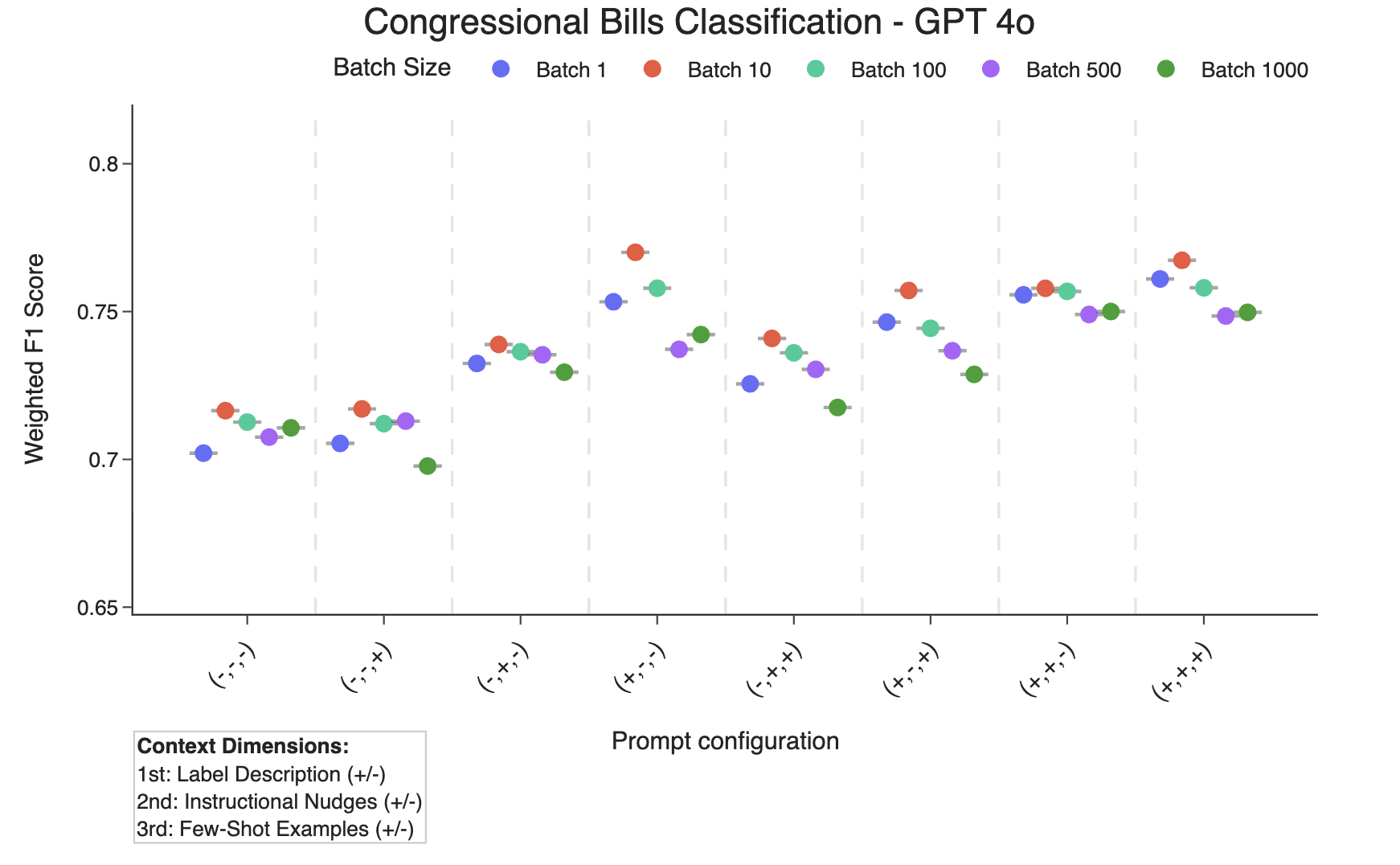}
\caption{Issue topic classification performance of GPT 4o across different contextual information configurations and input text batch sizes}
\end{figure}

\newpage
Accuracy performance of the two models across the numerous configurations presented in Figure 2 and Figure 3 exhibit a positive correlation with the complexity of the contextual information configuration. The leftmost configuration is the most minimal configuration where none of the three contextual information components are present. The next three configurations to the right of the most minimal configuration feature only one of the three components, and the next three configurations to the right of them feature two of the three components. The rightmost configuration has all three components. 

For each contextual information configuration, we present results with five different input text bath sizes. We tried a maximum batch size of 1000, where we started to observe significant decline in performance, especially with the GPT 4o model. Batch size 1 is what most researchers use when they use an instructed-tuned LLM for a task annotation task, but inserting multiple input texts within a single prompt might improve cost efficiency, but also enable some performance gains. The results from experiments with GPT 4o and Gemini 2.0 Flash indicate that presenting a single input text to annotate in a single LLM call will not give the most accurate output in more than half of the configurations.

\begin{figure}[H]
\centering
\includegraphics[width=0.95\linewidth]{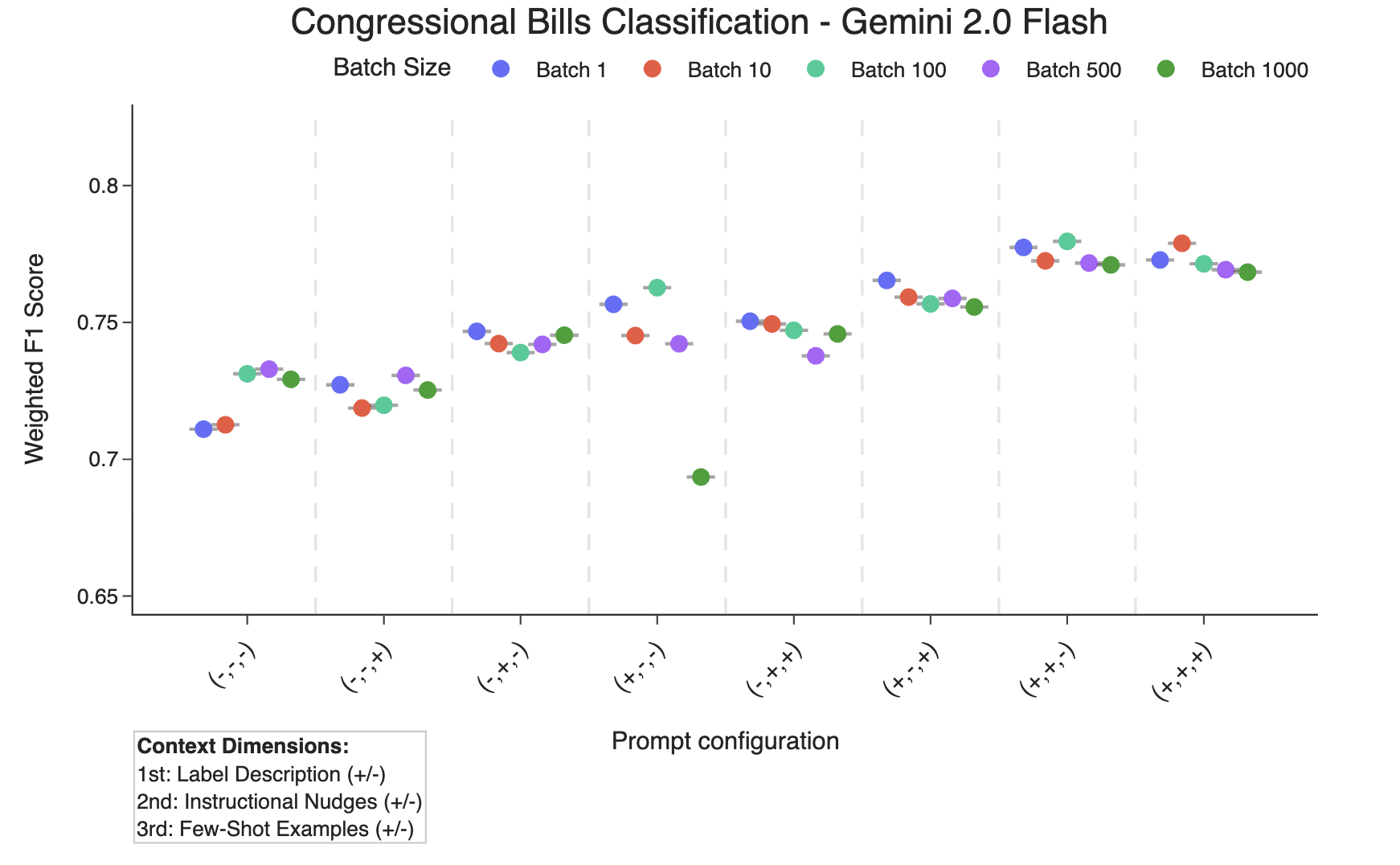}
\caption{Issue topic classification performance of Gemini 2.0 Flash across different contextual information configurations and input text batch sizes}
\end{figure}
\FloatBarrier

\subsection{Study 2 : Emotionality Detection in Open-Ended Survey Responses}

We now turn to our analysis of open-ended survey responses from the American National Election Studies (ANES) data. Before delving into emotionality in open-ended responses, it is important to note that the responses analyzed here were labelled after the cutoff dates of the language models used in this study, which ensures that these data could not have been part of the models’ training data and therefore potential concerns about data leakage or memorization. A large strand of research in political science has examined answers to the most important problem question in surveys, and their relevance for various micro- and macro-political phenomena. However, this literature almost exclusively focused on the frequency of MIP mentions. While the content of responses to “most important problem” (MIP) questions reveals public policy priorities, the emotional tone and affective language used by respondents (such as anger toward politicians, fear of economic collapse or war, or frustration with institutions) provide crucial insights into the intensity and valence of salience, shaping political mobilization and responsiveness beyond mere topic mention. Analyzing affective language in open-ended MIP answers uncovers how emotions amplify volatility (e.g., scandal-driven spikes) and signal deeper distrust, offering a richer measure of agenda dynamics than frequency coding alone. This dual focus on “what” and “how” enhances validity for studying anti-politics and issue thermostats.

In an attempt to demonstrate the utility of this approach, we present an illustrative case on the emotion-coding of open-ended answers to the MIP question in ANES surveys. For this illustrative case, we downloaded the open-ended answers to the ‘most important problem’ question that has been continuously asked in ANES surveys over the past decades.\footnote{The question takes the following form: ``What is the most important problem facing the country?'' Although slightly different versions of the question have been asked by various survey organizations since 1939, the question asked by ANES has followed the same structure across decades.} Respondents’ verbatim answers were recorded by interviewees, and, after any identifying information that might reveal the respondent’s identity was removed, verbatim answers were posted on the ANES website. Following Ekman’ (1990) framework of six basic emotions, namely, anger, disgust, fear, joy, sadness, and surprise, we coded these answers as emotional if they contained a word or phrase which was emotionally loaded (e.g., “love", “feel threatened", “hate”). Responses were coded as neutral if they described a problem without any expression of emotion, typically involving factual statements.

Below we provide two example answers for the emotional and neutral answer categories, the first being coded as emotional whereas the second as neutral:
\bigskip
\begin{quote}
``They should give the government back to the people. This country was supposed to be by the people, for the people. They should follow the Constitution. They're all liers, cheats, crooks don't care what the people want or say. They take away from us what we have worked hard for and earned get rid of the IRS and that bank that runs everything the federal bank that has 13 branches and runs everything. Stop messing with people let them farm or do whatever they want like we used to do without interference. No.''
\end{quote}
\bigskip
\begin{quote}
``The national debt. The money we owe out to other countries is tremendous. One time we have multi-trillion deficit in this country. We've loaned out money to other countries and they never paid us back.''
\end{quote}
\bigskip
In this illustrative case, we use the same prompt structure depicted in Figure 1, although the implementation of its individual components differs slightly from that in Section 3.1, given the nature of the task. In this application, we wrote the label descriptions rather than drawing from an existing codebook, as no established coding scheme exists for emotional and affective content in MIP responses. The instructional nudges consist of a single statement encouraging the model to approximate an expert-informed label distribution (see Appendix A), which reflects our substantive judgments about the relative prevalence of categories. The few-shot examples included in the prompt were selected separately from the randomly drawn analysis sample to avoid overlap. Finally, due to the smaller size of the MIP subsample, the maximum batch size used in this application was lower than in the bills study.

 \begin{figure}[H]
\centering
\includegraphics[width=0.95\linewidth]{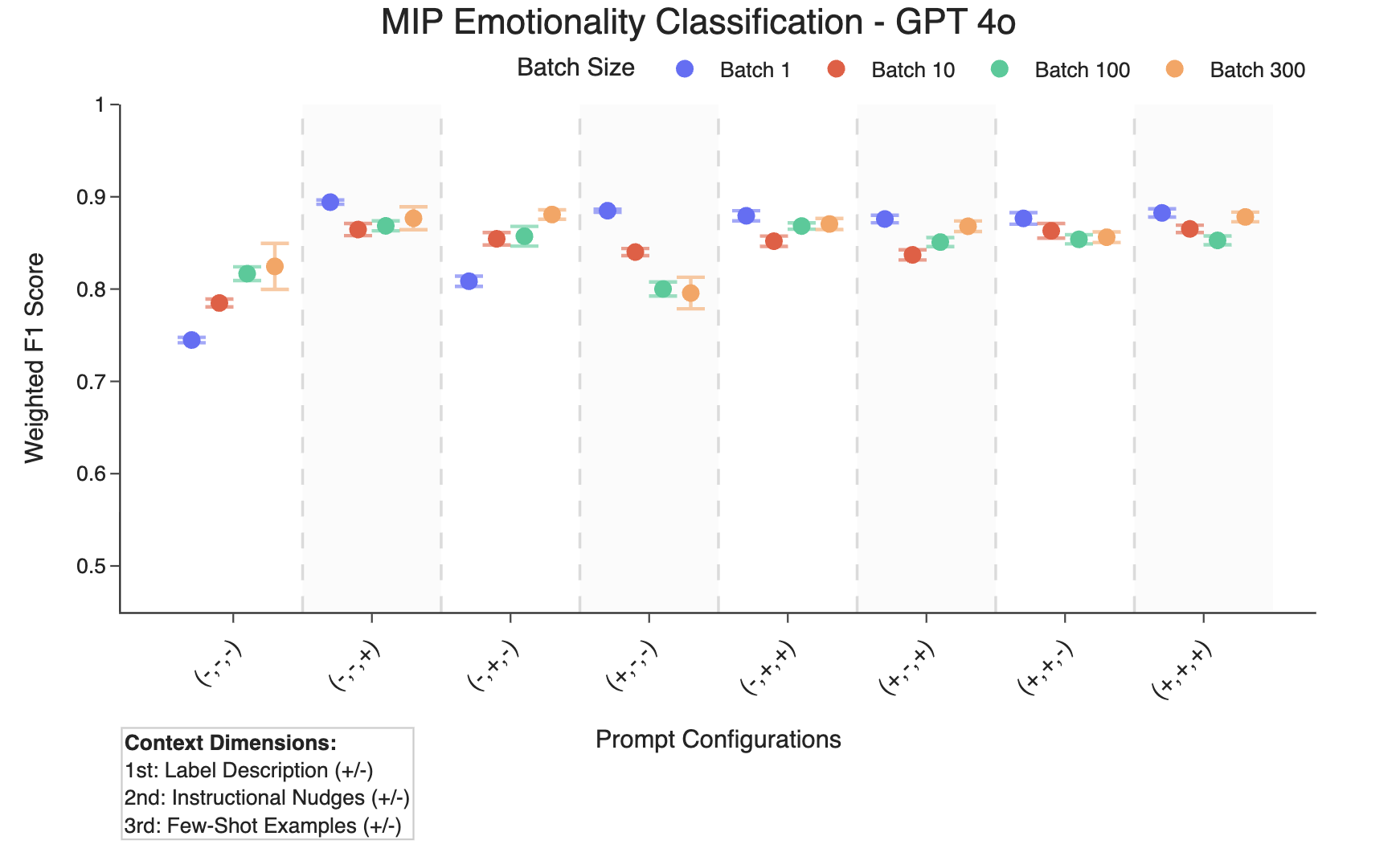}
\caption{Emotionality classification performance of GPT 4o across different contextual information configurations and input text batch sizes}
\end{figure}

In Figure 4 and Figure 5, we present the weighted F1 scores for MIP emotionality classification across different prompt configurations and batch sizes for GPT-4o and Gemini 2.0 Flash. For each prompt configuration, results are reported for four batch sizes (1, 10, 100, and 300), as was done in the previous section. Across both models, performance varies across prompt configurations defined by the inclusion or exclusion of label descriptions, instructional nudges, and few-shot examples. For GPT-4o, weighted F1 scores generally range between approximately 0.75 and 0.90 across configurations and batch sizes. For Gemini 2.0 Flash, performance spans a wider range, particularly at batch size 1, with higher batch sizes displaying relatively little variance across configurations. In both models, configurations that include additional contextual components tend to be associated with higher and more stable F1 scores across batch sizes.

\begin{figure}[H]
\centering
\includegraphics[width=0.95\linewidth]{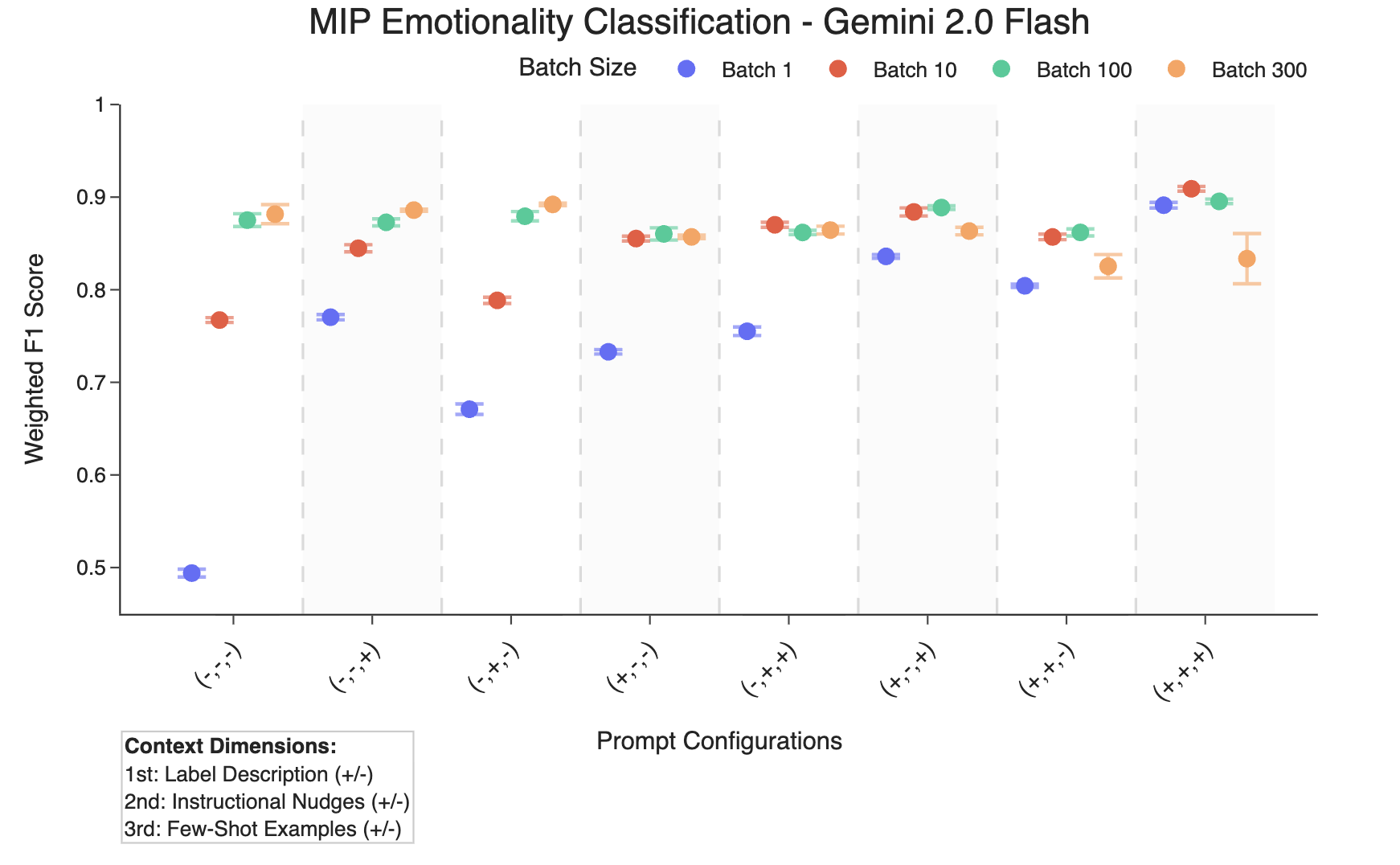}
\caption{Emotionality classification performance of Gemini 2.0 Flash across different contextual information configurations and input text batch sizes}
\end{figure}

\section{Discussion}

In this paper we presented two tests of LLM performance on classification tasks across varying configurations of context prompting and batch sizes. We aimed at opening the black box of LLM performance by systematically varying the input to assess the relative importance of context prompting elements and batch sizes, as well as testing how robust the patterns of performance were across multi and dual classification tasks and the possible inclusion of the data in the model’s training.

Several patterns of interest emerge. First, across all tests we see that prompting context matters in a pattern where prompt engineering tends to increase the F1-score. In much early work on text annotation using LLMs, researchers have been making use of minimalists prompts, which are likely a suboptimal strategy, and should be avoided in future research relying on LLMs. However, the difference between light prompt engineering and more heavy-handed prompt engineering (e.g making use of more strategies simultaneously) is oftentimes rather small, suggesting that the majority of the effect is achieved rather rapidly and then tapers off. It should be noted that while the performance gain in the later stages of prompt engineering is relatively modest, it is also practically easy to implement and monetarily cheap and thus most researchers should not feel limited in employing these techniques and reaping the advantages. It must be noted, that while we test a few strategies here, more are likely to emerge and we are unable to assess the effectiveness of adding these additional strategies along with ours based on current results.

There is also future research to be done in exploring the finer details of the individual prompt engineering strategies we employ here. For instance, future research could examine how many examples are needed to stabilize the results of few-shot prompting, and how this varies by models. Hopefully, as the field keeps advancing and models improve (Gunes and Florczak, 2025) many heterogeneities may slowly disappear and thus make this type of prompting strategy less important. However, the current heterogeneity may also be sparking the different recommendations currently found in the literature, where some authors actively discourage the use of LLMs for text annotation (Kristensen-McLachlan et. al., 2025). We show fairly good results on our tests, and therefore believe that LLMs can be used, but that it needs to be evaluated on a case-by-case basis.

We also observe heterogeneity between our two tasks and datasets. There can be several reasons for that. It could be the case that the labeling of emotion from text is a type of problem wherein the notion of a gold standard is itself challenged (Bisbee and Spirling, 2025) as is also the case with ground truth measurements (Gilardi, Alizadeh and Kubli, 2023; Törnberg, 2024). In any case, while many studies solely look at one type of texts to classify, the value added from comparing our rather diverse sets of texts and tasks is clear: it cannot be assumed that different tasks and texts follow a similar pattern in performance measures or engineering, and it cannot be assumed that performance measures necessarily decrease with the complexity of the task. While it is generally the case, performance is also dependent on the configuration of prompt engineering, batch size and model, as well as task and text type, and we therefore strongly encourage all users of LLM’s for text annotation tasks to conduct, report and evaluate validation efforts for their specific tasks rather than solely relying on existing research as validation. Therefore, documentation of the prompting strategy and the trial and error approach in each individual task is paramount, and we suggest that this should be included in appendix material. This is important for two reasons. First, our large-scale tests revealed that the temperature setting on LLMs does not necessarily guarantee deterministic (e.g. fully replicable) results even at a setting of 0 (See also: Yuan et. al., 2025). Second, we have instances of sudden, rather substantial, changes in classification performance following minor changes in prompting structure. This brittleness in model performance is an essential factor to take into account, and it can only really be discovered through multiple trials. Thus, while LLMs as text classification tools are clearly useful if the configuration is well chosen, researchers should be aware of the rather large range of possible outcomes if the configuration is not – and that eliminating all randomness in output is likely not feasible given the current models. This may be mitigated through local models, where deterministic parameters can be controlled.

As algorithms continue to improve it is also likely that the effects of prompt engineering may change. However, emerging research on the usefulness of LLM approaches in the face of rapid technological development paints varying pictures of how that may affect the trustworthiness of current results. For instance, Morucci and Spirling (2025) argue that the data used in political science is typically of a low dimensionality and that adding more complexity in terms of additional variables rarely provides us with much more accurate analyses than the usual low dimensionality structure. As such, the more advanced models will often be unable to significantly outperform the current ones. Should the same logic apply to LLM based classification, it may be the case that newer models may not exhibit significantly superior performance compared to those that are already available. However, the high degree of dimensionality in text data specifically may mean that improvements can still be made through more advanced methods in this particular use-case.

Future research should investigate the possibility of diminishing returns to prompt optimization in newer models compared to older. It may be the case that as models become better, their ability to accurately interpret varying instructions may also improve, and that this decreases the usefulness of prompt engineering. Procedures to automatically generate prompts is another research avenue that underscores that manual prompt-engineering is itself contested as an approach to optimizing LLM output. Iteratively generating prompt structures to maximize coding output by a fixed pipeline structure such as through DSPy (Khattab et. al., 2023) would likely improve the output while also keeping the costs of applying LLMs low. While the advantages are clear, this approach still implicitly accepts a black box approach to generating prompts, which should be kept in mind.

\section{Conclusion}

This work offers a structured approach to building context within a text classification prompt and to systematically testing how different contextual information components affect LLMs' accuracy and reproducibility in text classification tasks. Our experiments with GPT 4o and Gemini 2.0 Flash, spanning all possible combinations of the three contextual information components, showed that researchers should avoid prompt designs which contain only the minimum information needed for task execution, and should instead include richer contextual information along task-relevant dimensions. Our findings show large accuracy gains even when a single contextual information component is included in a text classification prompt, but accuracy does not always increase monotonically as the prompt becomes more complex with the addition of more than one contextual component. We also present evidence for the non-determinism of LLM responses even when model parameters are set to produce deterministic outputs, and our results show the extent of variability across different LLM requests with the exact same configuration also depends on prompt complexity.

In this work, we focused on a small number of models and contextual information components, and explored a limited range of variation along each of the tested components. Future research should explore our prompt engineering and experimentation strategy with a larger set of models, and especially with open source models which offer researchers greater control over data and conditions that affect reproducibility. We also need greater attention from researchers to experiment with different sizes of few-shot examples and different example selection strategies. By pursuing these lines of inquiry, researchers can better understand when context helps, when it saturates, and how to make LLM-based classification more dependable.

\clearpage
\newgeometry{margin=1in}
\singlespacing
\pagestyle{plain}
\pagenumbering{arabic}
\fontsize{10}{12}\selectfont
\setlength{\parindent}{15pt}
\setlength{\parskip}{0pt}
\setcounter{page}{22}
\thispagestyle{plain}
\begin{center}
{\LARGE\bfseries Supplementary Material\par}
\end{center}
\vspace{1.5em}

\section*{Appendix A: Prompt Examples}
\subsection*{Congressional Bills - Issue Topic Classification Prompt}
\begin{promptbox}{blue}{Label Descriptions}
\begin{verbatim}
Here is a list of policy issue topic labels (keys) and specific topics that
belong to that label category (values):

{'Macroeconomics': "Includes issues related to general domestic macroeconomic
policy and more specific issues including inflation, cost of living, prices,
interest rates, unemployment rate, impact of unemployment, training and
retraining, unemployment benefits, monetary policy, central bank, treasury,
public debt, budgeting, efforts to reduce deficits, tax policy, the impact of
taxes, and tax enforcement, manufacturing policy, industrial revitalization and
growth, wage or price control, emergency price controls.",
'Civil Rights': "Includes issues related to civil rights and minority rights.
More specifically, it includes issues related to minority, ethnic, and racial
group discrimination, sex, gender, and sexual discrimination, age
discrimination, mandatory retirement age policies, handicap and disease
discrimination, voting rights, expanding or contracting the franchise,
participation and related issues, freedom of speech, religious freedoms, other
types of freedom of expression, privacy rights, privacy of records, access to
government information, abortion rights and anti-government activity groups.",
'Health': "Includes issues related generally to health care, including
appropriations for general health care government agencies. More specifically,
it includes issues related to broad, comprehensive changes in the health care
system, health insurance reform, regulation, availability, and cost of
insurance, the regulation and promotion of pharmaceuticals, medical devices, and
clinical labs, facilities construction, regulation and payments, provider and
insurer payments and regulation, medical liability, malpractice issues, medical
fraud and abuse, and unfair practices, the supply and quantity of labor in the
health care industry, training and licensing, disease prevention and treatment,
health promotion, coverage and quality of infants and children care, school
health programs, mental health care, and mental health disease, long term care,
home health care, the terminally ill and rehabilitation services, prescription
drug coverage, programs to pay for prescription drugs, policy to reduce the cost
of prescription drugs, tobacco abuse, treatment, education and health effects,
alcohol and illegal drug abuse, treatment, education and health effects, health
care research and development.",
'Agriculture': "agriculture policy, the regulation and impact of agricultural
foreign trade, government subsidies to farmers and ranchers, food inspection and
safety, the regulation of agricultural marketing and providing information to
consumers regarding a healthy diet, animal and crop disease, pest control and
pesticide regulation, and welfare for domesticated animals, fishing, commercial
fishery regulation and conservation, agricultural research and development.",
'Labor': "employment, and pensions, worker safety and protection and
compensation for work-related injury and disease, job training for adult
workers, workforce development, and efforts to retrain displaced workers,
employee benefits, pensions, and retirement accounts, including
government-provided unemployment insurance, labor unions, collective bargaining,
employer-employee relations, fair labor standards, minimum wage and overtime
compensation, labor law, youth employment, child labor and job training for
youths, migrant, guest and seasonal workers.",
'Education': "education policy, higher education, student loans and education
finance, the regulation of colleges and universities, elementary and private
schools, school reform, safety in schools, efforts to generally improve
educations standards and outcomes, education of underprivileged students, adult
literacy programs, bilingual education needs, rural education initiatives,
vocational education for children and adults, special education for the
physically or mentally handicapped, education excellence, research and
development in education.",
'Environment': "environmental policy, domestic drinking water safety, supply,
pollution and additives, disposal and treatment of wastewater, solid water and
runoff, hazardous waste and toxic chemical regulation, treatment and disposal,
air pollution, climate change, noise pollution, recycling, reuse, resource
conservation, environmental hazards, indoor air contaminations, indoor hazardous
substances, forest protection, endangered species, control of the domestic
illicit trade in wildlife products, regulation of laboratory or performance
animals, land and water resource conservations, research and development in
environmental technology, not including alternative energy",
'Energy': "energy policy, nuclear energy, safety and security, and disposal of
nuclear waste, general electricity, hydropower, regulation of electric
utilities, natural gas and oil, drilling, oil spills and flaring, oil and gas
prices, shortages and gasoline regulation, coal production, use, trade, and
regulation, alternative and renewable energy, biofuels, hydrogen and geothermal
power, energy conservation and energy efficiency, energy research and
development.",
'Immigration': "immigration, refugees, and citizenship.",
'Transportation': "transportation policy, mass transportation construction,
regulation, safety and availability, highway construction, maintenance and
safety, air travel, regulation and safety of aviation, airports, air traffic
control, pilot training, aviation technology, railroads, rail travel, rail
freight, development and deployment of new rail technologies, maritime
transportation, inland waterways and channels, infrastructure and public works,
transportation research and development.",
'Law and Crime': "general law, crime and family issues, law enforcement
agencies, white collar crime, organized crime, counterfeiting and fraud,
cyber-crime, money laundering, illegal drug crime, criminal penalties for drug
crimes, court administration, judiciary appropriations, guidelines for bail,
pre-release, fines and legal representation, prisons and jails, parole systems,
juvenile crime and justice, juvenile prisons and jails, efforts to reduce
juvenile crime and recidivism, domestic violence, child welfare, family law,
domestic criminal and civil codes, control, prevention and impact of crime,
police and other general domestic security responses to terrorism.",
'Social Welfare': "social welfare policy, poverty assistance, food assistance
programs, programs to assess or alleviate welfare dependency and tax credits
directed at low-income families, elderly issues and elderly assistance,
government pensions, aid for people with physical or mental disabilities,
domestic volunteer associations, charities and youth organizations, parental
leave and childcare.",
'Housing': "housing and urban affairs, housing and community development,
neighborhood development, national housing policy, urban development and general
urban issues, rural housing, economic, infrastructure and other development in
non-urban areas, housing for low-income individuals and families, housing for
the elderly, housing for homeless, efforts to reduce homelessness, housing and
community development research and development.",
'Domestic Commerce': "government agencies regulating domestic commerce, the
regulation of national banking systems and other non-bank financial
institutions, the regulation and facilitation of securities and commodities
trading, regulation of investments and related industries and exchanges,
consumer finance, mortgages, credit cards, access to credit records, consumer
credit fraud, insurance regulation, fraud and abuse in the insurance industry,
the financial health of insurance industry, insurance availability and cost,
personal, commercial and municipal bankruptcies, corporate mergers, antitrust
regulation, corporate accounting and governance, corporate management, small
businesses, copyrights and patents, patent reform, intellectual property,
domestic natural disaster relief, disaster or flood insurance, natural disaster
preparedness, tourism regulation, promotion and impact, consumer fraud and
safety in domestic commerce, regulation and promotion of sports, gambling and
personal fitness, domestic commerce research and development.",
'Defense': "general defense policy, defense alliance and agreement, security
assistance, UN peacekeeping activities, military intelligence, espionage and
covert operations, military readiness, coordination of armed services air
support and sealift capabilities, national stockpiles of strategic materials,
nuclear weapons, nuclear proliferation, modernizations of nuclear equipment,
military aid to other countries, control of arms sales, military manpower,
military personnel and their dependents, military courts, general veterans'
issues, military procurement, conversion of old equipment, weapons and systems
evaluation, military installations, construction and land transfers, military
reserves and reserve affairs, military nuclear and hazardous waste disposal and
military environmental compliance, domestic civil defense, national security
responses to terrorism, non-contractor civilian personnel, civilian employment
in the defense industry, military base closings, military contractors,
war-related military operations, prisoners of war and collateral damage to
civilian populations, claims against the military, settlements for military
dependents, compensation for civilians injured in military operations, defense
research and development.",
'Technology': "space, science, technology and communications, the government use
of space and space resource exploitation agreements, government space programs
and space exploration, military use of space, regulation and promotion of
commercial use of space, commercial satellite technology, government efforts to
encourage commercial space development, science and technology transfer,
international science cooperation, telephones and telecommunication regulation,
infrastructure for high speed internet, mobile phones, broadcast industry,
published media, control of the electromagnetic spectrum, weather forecasting,
oceanography, geological surveys, weather forecasting research and technology,
computer industry, regulation of the internet, cyber security.",
'Foreign Trade': "negotiations, disputes and agreements, including tax treaties,
export regulation, subsidies, promotion and control, international private
business investment and corporate development, productivity of competitiveness
of domestic businesses and balance of payment issues, tariffs and other barriers
to imports, import regulation and impact of imports on domestic industries,
exchange rate and related issues.",
'International Affairs': "general international affairs and foreign aid, foreign
aid not directly targeting and increasing international development,
international resources exploitations and resources agreements, law of the sea
and international ocean conservation efforts, developing countries,
international finance and economic development banks, sovereign debt and
implications for international lending institutions, Western Europe and European
Union, foreign country or region, assessment of political issues in other
countries, relations between individual countries, human rights violations,
human rights treaties and conventions, UN reports on human rights, crimes
associated with genocide or crimes against humanity, international
organizations, NGOs, the United Nations, International Red Cross, UNESCO,
International Olympic Committee, International Criminal Court, international
terrorism, hijacking, acts of piracy in other countries, efforts to fight
international terrorism, international legal mechanisms to combat terrorism,
diplomats, diplomacy, embassies, citizens abroad, foreign diplomats in the
country, visas and passports.",
'Government Operations': government agencies, intergovernmental relations, local
government issues, general government efficiencies, bureaucratic oversight,
postal services, regulation of mail, post-civil service, government pensions,
general civil service issues, nominations and appointments, currency, national
mints, medals and commemorative coins, government procurement, government
contractors, contractor and procurement fraud, procurement processes and
systems, government property management, construction and regulation, tax
administration, enforcement, auditing for both individuals and corporations,
public scandal and impeachment, government branch relations, administrative
issues, constitutional reforms, regulation of political campaigns, campaign
finance, political advertising and voter registration, census and statistics
collection by government, capital city, claims against the government,
compensation for the victims of terrorist attacks, compensation policies without
other substantive provisions, national holidays and their observation.",
'Public Lands': "general public lands, water management, and territorial issues,
national parks, memorials, historic sites and recreation, the management and
staffing of cultural sites, indigenous affairs, indigenous lands, assistance to
indigenous people, natural resources, public lands, forest management, forest
fires, livestock grazing, water resources, water resource development and civil
works, flood control and research, territorial and dependency issues and
devolution.",
'Culture': "general cultural policy issues
'Private Bill: A private bill provides benefits to specified individuals
(including corporate bodies). Individuals sometimes request relief through
private legislation when administrative or legal remedies are exhausted. Many
private bills deal with immigration-granting citizenship or permanent residency.
Private bills may also be introduced for individuals who have claims against the
government, veterans' benefits claims, claims for military decorations, or
taxation problems. The title of a private bill usually begins with the phrase,
"For the relief of. . . ." if a private bill is passed in identical form by both
houses of Congress and is signed by the president, it becomes a private law."}
"""
\end{verbatim}
\end{promptbox}

\begin{promptbox}{orange}{Instructional Nudges}
\begin{verbatim}
I will show you congressional bill titles. Please assign a label to each
title. Give me only the title number and the label, and put one label per line.
Use ': ' to separate numbers from labels. Don't use any label which is not
listed among the candidate labels.
Here are some notes:

                    - The category defense also includes issues related to
                    veterans affairs and welfare benefits to current military
                    members,

                    - The category public lands encompasses issues including,
                    but not limited to, Native Americans affairs, national parks
                    and national forests and interstate highways,

                    - Abortion related issues belong to the category of civil
                    rights,

                    - Banking and finance related issues are related to domestic
                    commerce category,

                    - Issues related countries other than the United States are
                    related to the category of international affairs,

                    - Technology category also includes issues related to
                    science and space,

                    - Titles which contain keywords such as import, export,
                    tariff and duty are related to the category of foreign
                    trade,

                    - Issues related to American state governments and the
                    federal government organizations and their employees, not
                    including the members of the army, are related to government
                    operations category,

                    - Culture category is very rare, less than one percent, in
                    the congressional bill title data
\end{verbatim}
\end{promptbox}

\begin{promptbox}{green}{Few-Shot Examples}
\begin{verbatim}
Here are a few example titles that belong to the "macroeconomics" category:

                        1: 'To amend the Congressional Budget Act of 1974 to
                        extend and reduce the discretionary spending limits and
                        to extend
                        the pay-as-you-go requirements set forth in the Balanced
                        Budget and Emergency Deficit Control Act of 1985.'

                        2: 'To amend the Balanced Budget and Emergency Deficit
                        Control Act of 1985 to modify the discretionary spending
                        limits
                        to take into account savings resulting from the
                        reduction in the number of Federal employees.'

Here are a few example titles that belong to the "civil
                    rights" category:

                        1: 'A bill to direct the United States Sentencing
                        Commission to make sentencing guidelines for Federal
                        criminal
                        cases that provide sentencing enhancements for hate
                        crimes.'

                        2: 'A bill to provide for unbiased consideration of
                        applicants to medical schools.'

                        3: 'To prohibit discrimination on account of sex in the
                        payment of wages by employers engaged in commerce or
                        in operation of industries affecting commerce, and to
                        provide procedures for assisting employees in collecting
                        wages lost by reason of any such discrimination'

Here are a few example titles that belong to the "health"
                    category:

                        1: 'A bill to provide grants to the States to assist
                        them in informing and educating children in schools with
                        respect to the harmful effects of. tobacco, alcohol, and
                        other potentially deleterious consumables.'

                        2: 'To amend title XXVII of the Public Health Service
                        Act and title I of the Employee Retirement Income
                        Security Act of 1974 to require that group and
                        individual health insurance coverage and group health
                        plans
                        provide comprehensive coverage for childhood
                        immunization.'

Here are a few example titles that belong to the
                    "agriculture" category:

                        1: 'To impose import limitations on certain meat and
                        meat products'

                        2: 'To enable wheat producers, processors, and
                        end-product manufacturers of wheat foods to work
                        together to
                        establish, finance, and administer a coordinated program
                        of research, education, and promotion to maintain
                        and expand markets for wheat and wheat products for use
                        as human foods within the United States'

Here are a few example title that belong to the "labor"
                    category:

                        1: 'To improve the safety conditions of persons working
                        in the coal mining industry of the United States'

                        2: 'A bill to improve the H-2A agricultural worker
                        program for use by dairy workers, sheepherders, and
                        goat herders, and for other purposes.'

Here are a few example titles that belong to the
                    "education" category:

                        1: 'A bill to amend the Internal Revenue Code of 1954 to
                        exclude from gross income the amount of certain
                        cancellations of indebtedness under student loan
                        programs.'

                        2: 'To establish a grant program for nebulizers in
                        elementary and secondary schools.'

Here are a few example titles that belong to the
                    "environment" category:

                        1: 'To provide for the elimination of the use of lead in
                        motor vehicle fuel and the installation of
                        adequate antipollution devices on motor vehicles, and
                        for other purposes'

                        2: 'To amend the Lacey Act Amendments of 1981 to clarify
                        provisions enacted by the Captive Wildlife Safety Act,
                        to further the conservation of certain wildlife species,
                        and for other purposes.'

Here are a few example titles that belong to the "energy"
                    category:

                        1: 'To enhance and improve the energy security of the
                        United States, expand economic development,
                        increase agricultural income, and improve environmental
                        quality by reauthorizing and improving
                        the renewable energy systems and energy efficiency
                        improvements program of the Department of Agriculture
                        through fiscal year 2012, and for other purposes.'

                        2: 'A bill to amend Public Law 95-209 to increase the
                        authorization for appropriations to the Nuclear
                        Regulatory Commission in accordance with Section 261 of
                        the Atomic Energy Act of 1954, as amended,
                        and Section 305 of the Energy Reorganization Act of
                        1974, as amended, and for other purposes.'

Here are a few example titles that belong to the
                    "immigration" category:

                        1: 'A bill to authorize the admission into the United
                        States, under a quota for Koreans, persons of
                        the Korean race, to make them racially eligible for
                        naturalization, and for other purposes'

                        2: 'A bill to improve agricultural job opportunities,
                        benefits, and security for aliens in the
                        United States and for other purposes.'

Here are a few example titles that belong to the
                    "transportation" category:

                        1: 'A bill to amend the Act of July 26, 1956, to give
                        the Muscatine Bridge Commission additional time
                        to construct, maintain, and operate a bridge across the
                        Mississippi River at or near the city of Muscatine,
                        Iowa, and the town of Drury, Ill'

                        2: 'To amend title 49, United States Code, to provide
                        relief to the airline industry, to reform the
                        Federal Aviation Administration, and to make technical
                        corrections, and for other purposes.'

Here are a few example titles that belong to the "law and
                    crime" category:

                        1: 'To assist in reducing crime by requiring speedy
                        trials in cases of persons charged with violations of
                        Federal criminal laws, to strengthen controls over
                        dangerous defendants released prior to trial, to provide
                        means for effective supervision and control of such
                        defendants, and for other purposes'

                        2: 'A bill to amend title 18, United States Code, to
                        exempt qualified current and former law enforcement
                        officers from State laws prohibiting the carrying of
                        concealed handguns.'

Here are a few example titles that belong to the "social
                    welfare" category:

                        1: 'To repeal those provisions of law which exclude from
                        the Federal old-age and survivors insurance system
                        service performed by an individual in the em ploy of his
                        son, daughter, spouse, or parent'

                        2: 'To amend the Internal Revenue Code of 1986 to repeal
                        the 1993 increase in taxes on Social Security benefits.'

Here are a few example titles that belong to the "housing"
                    category:

                        1: 'To extend and liberalize the direct home loan
                        program for veterans, to extend the guaranteed home loan
                        program, to provide special assistance to paraplegic
                        veterans under the direct home loan program, to
                        stimulate
                        the making of direct farm housing loans, '

                        2: 'To reform the regulation of certain housing-related
                        Government-sponsored enterprises.'

Here are a few example titles that belong to the "domestic
                    commerce" category:

                        1: 'To amend the Internal Revenue Code of 1954 to
                        provide that the term purchase for purposes of section
                        334 (b) (2)
                        is to include certain indirect purchases of stock
                        through the purchase of the stock of another
                        corporation'

                        2: 'To amend the Internal Revenue Code of 1986 to
                        permanently modify the limitations on the deduction of
                        interest by
                        financial institutions which hold tax-exempt bonds, and
                        for other purposes.'

Here are a few example titles that belong to the "defense"
                    category:

                        1: 'A bill to establish requirements for notification of
                        Congress before the closure of, or significant reduction
                        in force at, any military installation is carried out.'

                        2: 'To deauthorize the Military Selective Service Act,
                        including the registration requirement and the
                        activities
                        of civilian local boards, civilian appeal boards, and
                        similar local agencies of the Selective Service System,
                        except
                        during a national emergency declared by the President,
                        and for other purposes.'

                    - Here are a few titles that belong to the "technology"
                    category:

                        1: 'To direct the National Aeronautics and Space
                        Administration to plan to return to the Moon and develop
                        a sustained
                        human presence on the Moon.'

                        2: 'A bill to amend the Internal Revenue Code of 1954 to
                        provide incentives for research and development by
                        providing
                        an increased investment credit or the allowance of rapid
                        amortization, and for other purposes.'

Here are a few example titles that belong to the "foreign
                    trade" category:

                        1: 'To suspend temporarily the duty on Compound T3028.'

                        2: 'A bill to amend the Tariff Schedules of the United
                        States to permit the importation of upholstery
                        regulators,
                        upholsterer's regulating needles, and upholsterer's pins
                        free of duty.'

Here are a few example titles that belong to the
                    "international affairs" category:

                        1: 'To amend the Mutual Security Act of 1951 to provide
                        for the termination of assistance to any nation which
                        does not
                        make a full contribution to the development and
                        maintenance of the defensive strength of the free world'

                        2: 'To clarify the application of section 304 of the
                        Tariff Act of 1930 as it relates to articles from areas
                        of
                        the West Bank and Gaza that are not administered by
                        Israel.'

Here are a few example titles that belong to the
                    "government operations" category:

                        1: 'To retain coverage under the laws providing employee
                        benefits, such as compensation for injury, retirement,
                        life insurance, and health benefits for employees of the
                        Government of the United States who transfer to
                        Indian tribal organizations to perform services in
                        connection with governmental or other activities
                        which are or have been performed by Government employees
                        in or for Indian communities, and for other purposes',

                        2: 'A bill to amend the Clinger-Cohen Act of 1996 to
                        provide individual federal agencies and the executive
                        branch as
                        a whole with increased incentives to use the
                        share-in-savings program under that Act, to ease the use
                        of such program,
                        and for other purposes.'

Here are a few example titles that belong to the "public
                    lands" category:

                        1: 'To provide for the establishment of the Clara Barton
                        House National Historic Site in the State of Maryland,
                        and for other purposes'

                        2: 'To provide for the transfer of certain real property
                        to the National Park Service for inclusion in the
                        Upper Delaware Scenic and Recreational River.'

Here are a few example bill titles that belong to the
                    "private bill" category:

                        1: 'A bill for the relief of Andrew D, Sumner.'

                        2: 'A bill for the relief of Felix Hernandez-Arana and
                        his wife, Felicia Ogaldez-De Hernandez.
\end{verbatim}
\end{promptbox}

\subsection*{ANES (MIP) - Emotionality Classification Prompt}
Below is a list of open-ended survey responses. Classify each of them into one
of the following categories: neutral or emotional.
\begin{promptbox}{blue}{Label Descriptions}
\begin{verbatim}
A statement is emotional if it contains a word or phrase which is emotionally
loaded, such as "love", "feel threatened" etc. It is neutral if there is just a
problem description or factual statement.
\end{verbatim}
\end{promptbox}

\begin{promptbox}{orange}{Instructional Nudges}
\begin{verbatim}
Note: We manually labelled a similar sized random sample before, and we estimate
that the share of "neutral" labels must be around 75 percent. So, you should aim
to achive a similar distribution in your predicted labels.
\end{verbatim}
\end{promptbox}

\begin{promptbox}{green}{Few-Shot Examples}
\begin{verbatim}
Here are some example responses for each category:

Neutral

- "race relations, economy, and wealth gap",
- "gas prices, they are going up!!"
- "Cuts to military members and their families and their benefits."

Emotional

- "Homeless people -- I know there are a lot of homelessness. Our water
situation -- it's bad. Pollution -- is bad."
- "the constant lies. what is the truth!? why is the media wanting to create a
war!"
- "They should give the government back to the people.  This  country was
supposed to be by the people, for the people.  They should follow the
Constitution.  They're all liers,  cheats, crooks  don't care what the people
want or say.  They take away from us what we have worked hard for and  earned
get rid of the IRS and that bank that runs  everything  the federal bank that
has 13 branches  and runs everything.  Stop messing with people  let them  farm
or do whatever they want like we used to do without  interference.  No."
- "Raping  those poor women!"
- "The lack of love because if we had that everything else would disappear."
- "The most, well, the arms race I'm very concerned about it. The terrorism
problem  we're going to have to face it in our own back yard.  In Texas they'll
come straight up from Mexico or Nicaragua Immigration  illegals coming and we
must do something about it.  no.
\end{verbatim}
\end{promptbox}
\clearpage
\setcounter{table}{0}
\renewcommand{\thetable}{B\arabic{table}}
\section*{Appendix B: Summary Tables}
\subsection*{Congressional Bills - Issue Topic Classification}
\begin{longtable}{lrrrrrr}
\caption{Congressional Bills - Issue Topic Classification Performance Metrics (GPT 4o)}\label{tab:bills-summary-gpt4o}\\
\toprule
Label Desc. & Inst. Nudges & Few-Shot & Batch Size & Accuracy & F1 & Weighted F1 \\
\midrule
\endfirsthead
\caption[]{Congressional Bills - Issue Topic Classification Performance Metrics (GPT 4o) (continued)}\\
\toprule
Label Desc. & Inst. Nudges & Few-Shot & Batch Size & Accuracy & F1 & Weighted F1 \\
\midrule
\endhead
\midrule
\multicolumn{7}{r}{\textit{Continued on next page}}\\
\midrule
\endfoot
\bottomrule
\endlastfoot

- & - & - & 1 & 0.689 & 0.613 & 0.702 \\
- & - & - & 10 & 0.710 & 0.646 & 0.716 \\
- & - & - & 100 & 0.706 & 0.614 & 0.713 \\
- & - & - & 500 & 0.702 & 0.605 & 0.708 \\
- & - & - & 1000 & 0.705 & 0.607 & 0.711 \\
- & - & + & 1 & 0.697 & 0.608 & 0.705 \\
- & - & + & 10 & 0.712 & 0.620 & 0.717 \\
- & - & + & 100 & 0.705 & 0.617 & 0.712 \\
\end{longtable}

\clearpage
\subsection*{Congressional Bills - Issue Topic Classification}
\begin{longtable}{lrrrrrr}
\caption{Congressional Bills - Issue Topic Classification Performance Metrics (Gemini 2.0 Flash)}\label{tab:bills-summary-gemini2flash}\\
\toprule
Label Desc. & Inst. Nudges & Few-Shot & Batch Size & Accuracy & F1 & Weighted F1 \\
\midrule
\endfirsthead
\caption[]{Congressional Bills - Issue Topic Classification Performance Metrics (Gemini 2.0 Flash) (continued)}\\
\toprule
Label Desc. & Inst. Nudges & Few-Shot & Batch Size & Accuracy & F1 & Weighted F1 \\
\midrule
\endhead
\midrule
\multicolumn{7}{r}{\textit{Continued on next page}}\\
\midrule
\endfoot
\bottomrule
\endlastfoot

- & - & - & 1 & 0.711 & 0.618 & 0.711 \\
- & - & - & 10 & 0.709 & 0.621 & 0.713 \\
- & - & - & 100 & 0.726 & 0.665 & 0.731 \\
- & - & - & 500 & 0.726 & 0.665 & 0.733 \\
- & - & - & 1000 & 0.722 & 0.663 & 0.729 \\
- & - & + & 1 & 0.726 & 0.660 & 0.727 \\
- & - & + & 10 & 0.717 & 0.656 & 0.719 \\
- & - & + & 100 & 0.715 & 0.652 & 0.720 \\
\end{longtable}

\clearpage
\subsection*{ANES (MIP) - Emotionality Classification}
\begin{longtable}{lrrrrrr}
\caption{ANES (MIP) - Emotionality Classification Performance Metrics (Gemini 2.0 Flash)}\label{tab:mip-summary-gemini}\\
\toprule
Label Desc. & Inst. Nudges & Few-Shot & Batch Size & Accuracy & F1 & Weighted F1 \\
\midrule
\endfirsthead
\caption[]{ANES (MIP) - Emotionality Classification Performance Metrics (Gemini 2.0 Flash) (continued)}\\
\toprule
Label Desc. & Inst. Nudges & Few-Shot & Batch Size & Accuracy & F1 & Weighted F1 \\
\midrule
\endhead
\midrule
\multicolumn{7}{r}{\textit{Continued on next page}}\\
\midrule
\endfoot
\bottomrule
\endlastfoot

- & - & - & 1 & 0.486 & 0.485 & 0.494 \\
- & - & - & 10 & 0.749 & 0.722 & 0.767 \\
- & - & - & 100 & 0.870 & 0.836 & 0.875 \\
- & - & - & 300 & 0.877 & 0.844 & 0.882 \\
- & - & + & 1 & 0.752 & 0.728 & 0.770 \\
- & - & + & 10 & 0.836 & 0.800 & 0.845 \\
- & - & + & 100 & 0.866 & 0.834 & 0.873 \\
- & - & + & 300 & 0.882 & 0.848 & 0.886 \\
- & + & - & 1 & 0.647 & 0.634 & 0.671 \\
- & + & - & 10 & 0.772 & 0.742 & 0.789 \\
- & + & - & 100 & 0.874 & 0.841 & 0.879 \\
- & + & - & 300 & 0.888 & 0.856 & 0.892 \\
+ & - & - & 1 & 0.711 & 0.690 & 0.733 \\
+ & - & - & 10 & 0.849 & 0.810 & 0.855 \\
+ & - & - & 100 & 0.867 & 0.798 & 0.860 \\
+ & - & - & 300 & 0.868 & 0.789 & 0.857 \\
- & + & + & 1 & 0.735 & 0.710 & 0.755 \\
- & + & + & 10 & 0.865 & 0.827 & 0.870 \\
- & + & + & 100 & 0.856 & 0.817 & 0.862 \\
- & + & + & 300 & 0.858 & 0.823 & 0.864 \\
+ & - & + & 1 & 0.825 & 0.794 & 0.836 \\
+ & - & + & 10 & 0.881 & 0.844 & 0.884 \\
+ & - & + & 100 & 0.887 & 0.849 & 0.889 \\
+ & - & + & 300 & 0.874 & 0.798 & 0.863 \\
+ & + & - & 1 & 0.790 & 0.758 & 0.804 \\
+ & + & - & 10 & 0.851 & 0.812 & 0.857 \\
+ & + & - & 100 & 0.869 & 0.800 & 0.862 \\
+ & + & - & 300 & 0.845 & 0.736 & 0.825 \\
+ & + & + & 1 & 0.888 & 0.854 & 0.891 \\
+ & + & + & 10 & 0.908 & 0.875 & 0.909 \\
+ & + & + & 100 & 0.896 & 0.854 & 0.895 \\
+ & + & + & 300 & 0.854 & 0.747 & 0.834 \\
\end{longtable}

\clearpage
\subsection*{ANES (MIP) - Emotionality Classification}
\begin{longtable}{lrrrrrr}
\caption{ANES (MIP) - Emotionality Classification Performance Metrics (GPT 4o)}\label{tab:mip-summary-gpt}\\
\toprule
Label Desc. & Inst. Nudges & Few-Shot & Batch Size & Accuracy & F1 & Weighted F1 \\
\midrule
\endfirsthead
\caption[]{ANES (MIP) - Emotionality Classification Performance Metrics (GPT 4o) (continued)}\\
\toprule
Label Desc. & Inst. Nudges & Few-Shot & Batch Size & Accuracy & F1 & Weighted F1 \\
\midrule
\endhead
\midrule
\multicolumn{7}{r}{\textit{Continued on next page}}\\
\midrule
\endfoot
\bottomrule
\endlastfoot

- & - & - & 1 & 0.724 & 0.700 & 0.745 \\
- & - & - & 10 & 0.768 & 0.743 & 0.785 \\
- & - & - & 100 & 0.803 & 0.773 & 0.817 \\
- & - & - & 300 & 0.813 & 0.780 & 0.825 \\
- & - & + & 1 & 0.893 & 0.856 & 0.894 \\
- & - & + & 10 & 0.859 & 0.821 & 0.865 \\
- & - & + & 100 & 0.865 & 0.824 & 0.869 \\
- & - & + & 300 & 0.874 & 0.833 & 0.877 \\
- & + & - & 1 & 0.794 & 0.764 & 0.809 \\
- & + & - & 10 & 0.848 & 0.809 & 0.854 \\
- & + & - & 100 & 0.850 & 0.814 & 0.857 \\
- & + & - & 300 & 0.877 & 0.841 & 0.881 \\
+ & - & - & 1 & 0.881 & 0.847 & 0.885 \\
+ & - & - & 10 & 0.831 & 0.796 & 0.840 \\
+ & - & - & 100 & 0.784 & 0.759 & 0.800 \\
+ & - & - & 300 & 0.780 & 0.754 & 0.796 \\
- & + & + & 1 & 0.885 & 0.826 & 0.879 \\
- & + & + & 10 & 0.854 & 0.793 & 0.852 \\
- & + & + & 100 & 0.874 & 0.810 & 0.868 \\
- & + & + & 300 & 0.874 & 0.816 & 0.871 \\
+ & - & + & 1 & 0.873 & 0.833 & 0.876 \\
+ & - & + & 10 & 0.828 & 0.790 & 0.837 \\
+ & - & + & 100 & 0.844 & 0.806 & 0.851 \\
+ & - & + & 300 & 0.864 & 0.824 & 0.868 \\
+ & + & - & 1 & 0.873 & 0.835 & 0.877 \\
+ & + & - & 10 & 0.859 & 0.817 & 0.863 \\
+ & + & - & 100 & 0.852 & 0.801 & 0.854 \\
+ & + & - & 300 & 0.850 & 0.811 & 0.856 \\
+ & + & + & 1 & 0.886 & 0.833 & 0.883 \\
+ & + & + & 10 & 0.863 & 0.816 & 0.865 \\
+ & + & + & 100 & 0.853 & 0.795 & 0.853 \\
+ & + & + & 300 & 0.877 & 0.833 & 0.878 \\
\end{longtable}

\clearpage
\setcounter{table}{0}
\renewcommand{\thetable}{C\arabic{table}}
\section*{Appendix C: Class Specific Tables}
\subsection*{Congressional Bills - Issue Topic Classification (-,-,-)}
\begin{table}[H]
\centering
\setlength{\tabcolsep}{3pt}
\small
\caption{Congressional Bills - Issue Topic Classification Performance Metrics (GPT 4o) - (-,-,-)}
\begin{sideways}
\begin{tabular}{|l|c|c|c|c|c|c|c|c|c|c|c|c|c|c|c|}
\hline
Class & Precision & Recall & F1 & Precision & Recall & F1 & Precision & Recall & F1 & Precision & Recall & F1 & Precision & Recall & F1 \\
 & \multicolumn{3}{c|}{1} & \multicolumn{3}{c|}{10} & \multicolumn{3}{c|}{100} & \multicolumn{3}{c|}{500} & \multicolumn{3}{c|}{1000} \\
\hline
1 & 0.493 & 0.709 & 0.582 & 0.509 & 0.704 & 0.591 & 0.504 & 0.582 & 0.540 & 0.451 & 0.582 & 0.508 & 0.411 & 0.745 & 0.530 \\
2 & 0.643 & 0.479 & 0.549 & 0.652 & 0.479 & 0.552 & 0.641 & 0.436 & 0.519 & 0.771 & 0.394 & 0.521 & 0.884 & 0.404 & 0.555 \\
3 & 0.716 & 0.888 & 0.793 & 0.743 & 0.871 & 0.802 & 0.732 & 0.881 & 0.800 & 0.749 & 0.841 & 0.792 & 0.760 & 0.837 & 0.797 \\
4 & 0.763 & 0.817 & 0.789 & 0.745 & 0.787 & 0.765 & 0.773 & 0.777 & 0.775 & 0.754 & 0.761 & 0.758 & 0.757 & 0.726 & 0.741 \\
5 & 0.532 & 0.649 & 0.584 & 0.549 & 0.673 & 0.604 & 0.515 & 0.631 & 0.567 & 0.527 & 0.571 & 0.549 & 0.438 & 0.565 & 0.494 \\
6 & 0.728 & 0.843 & 0.781 & 0.720 & 0.843 & 0.777 & 0.718 & 0.786 & 0.751 & 0.723 & 0.786 & 0.753 & 0.678 & 0.730 & 0.703 \\
\hline
\end{tabular}
\end{sideways}
\end{table}
\clearpage

\subsection*{Congressional Bills - Issue Topic Classification (-,-,+)}
\begin{table}[H]
\centering
\setlength{\tabcolsep}{3pt}
\small
\caption{Congressional Bills - Issue Topic Classification Performance Metrics (GPT 4o) - (-,-,+)}
\begin{sideways}
\begin{tabular}{|l|c|c|c|c|c|c|c|c|c|c|c|c|c|c|c|}
\hline
Class & Precision & Recall & F1 & Precision & Recall & F1 & Precision & Recall & F1 & Precision & Recall & F1 & Precision & Recall & F1 \\
 & \multicolumn{3}{c|}{1} & \multicolumn{3}{c|}{10} & \multicolumn{3}{c|}{100} & \multicolumn{3}{c|}{500} & \multicolumn{3}{c|}{1000} \\
\hline
1 & 0.554 & 0.393 & 0.460 & 0.551 & 0.413 & 0.472 & 0.564 & 0.383 & 0.456 & 0.548 & 0.347 & 0.425 & 0.738 & 0.158 & 0.261 \\
2 & 0.578 & 0.511 & 0.542 & 0.662 & 0.543 & 0.596 & 0.746 & 0.500 & 0.599 & 0.759 & 0.436 & 0.554 & 0.848 & 0.415 & 0.557 \\
3 & 0.727 & 0.892 & 0.801 & 0.743 & 0.861 & 0.797 & 0.756 & 0.871 & 0.809 & 0.771 & 0.854 & 0.810 & 0.749 & 0.851 & 0.797 \\
4 & 0.807 & 0.766 & 0.786 & 0.819 & 0.782 & 0.800 & 0.800 & 0.772 & 0.786 & 0.783 & 0.751 & 0.767 & 0.807 & 0.721 & 0.761 \\
5 & 0.618 & 0.560 & 0.588 & 0.664 & 0.589 & 0.625 & 0.589 & 0.554 & 0.571 & 0.573 & 0.536 & 0.554 & 0.611 & 0.524 & 0.564 \\
6 & 0.749 & 0.843 & 0.793 & 0.725 & 0.811 & 0.766 & 0.757 & 0.805 & 0.780 & 0.694 & 0.786 & 0.737 & 0.690 & 0.799 & 0.741 \\
\hline
\end{tabular}
\end{sideways}
\end{table}
\clearpage

\subsection*{Congressional Bills - Issue Topic Classification (-,+,-)}
\begin{table}[H]
\centering
\setlength{\tabcolsep}{3pt}
\small
\caption{Congressional Bills - Issue Topic Classification Performance Metrics (GPT 4o) - (-,+,-)}
\begin{sideways}
\begin{tabular}{|l|c|c|c|c|c|c|c|c|c|c|c|c|c|c|c|}
\hline
Class & Precision & Recall & F1 & Precision & Recall & F1 & Precision & Recall & F1 & Precision & Recall & F1 & Precision & Recall & F1 \\
 & \multicolumn{3}{c|}{1} & \multicolumn{3}{c|}{10} & \multicolumn{3}{c|}{100} & \multicolumn{3}{c|}{500} & \multicolumn{3}{c|}{1000} \\
\hline
1 & 0.625 & 0.459 & 0.529 & 0.726 & 0.393 & 0.510 & 0.531 & 0.347 & 0.420 & 0.526 & 0.367 & 0.432 & 0.488 & 0.413 & 0.448 \\
2 & 0.722 & 0.553 & 0.627 & 0.743 & 0.553 & 0.634 & 0.778 & 0.521 & 0.624 & 0.847 & 0.532 & 0.654 & 0.894 & 0.447 & 0.596 \\
3 & 0.785 & 0.803 & 0.794 & 0.774 & 0.800 & 0.787 & 0.775 & 0.827 & 0.800 & 0.789 & 0.786 & 0.788 & 0.792 & 0.773 & 0.782 \\
4 & 0.773 & 0.726 & 0.749 & 0.771 & 0.701 & 0.734 & 0.779 & 0.751 & 0.765 & 0.761 & 0.711 & 0.735 & 0.770 & 0.680 & 0.722 \\
5 & 0.746 & 0.595 & 0.662 & 0.718 & 0.667 & 0.691 & 0.704 & 0.595 & 0.645 & 0.685 & 0.530 & 0.597 & 0.679 & 0.542 & 0.603 \\
6 & 0.781 & 0.717 & 0.748 & 0.786 & 0.717 & 0.750 & 0.748 & 0.692 & 0.719 & 0.748 & 0.692 & 0.719 & 0.723 & 0.673 & 0.697 \\
\hline
\end{tabular}
\end{sideways}
\end{table}
\clearpage

\subsection*{Congressional Bills - Issue Topic Classification (+,-,-)}
\begin{table}[H]
\centering
\setlength{\tabcolsep}{3pt}
\small
\caption{Congressional Bills - Issue Topic Classification Performance Metrics (GPT 4o) - (+,-,-)}
\begin{sideways}
\begin{tabular}{|l|c|c|c|c|c|c|c|c|c|c|c|c|c|c|c|}
\hline
Class & Precision & Recall & F1 & Precision & Recall & F1 & Precision & Recall & F1 & Precision & Recall & F1 & Precision & Recall & F1 \\
 & \multicolumn{3}{c|}{1} & \multicolumn{3}{c|}{10} & \multicolumn{3}{c|}{100} & \multicolumn{3}{c|}{500} & \multicolumn{3}{c|}{1000} \\
\hline
1 & 0.567 & 0.648 & 0.605 & 0.539 & 0.663 & 0.595 & 0.437 & 0.704 & 0.539 & 0.436 & 0.592 & 0.502 & 0.426 & 0.796 & 0.555 \\
2 & 0.833 & 0.638 & 0.723 & 0.762 & 0.649 & 0.701 & 0.758 & 0.500 & 0.603 & 0.745 & 0.436 & 0.550 & 0.829 & 0.362 & 0.504 \\
3 & 0.789 & 0.864 & 0.825 & 0.782 & 0.841 & 0.810 & 0.762 & 0.858 & 0.807 & 0.767 & 0.814 & 0.789 & 0.767 & 0.837 & 0.801 \\
4 & 0.773 & 0.863 & 0.815 & 0.815 & 0.848 & 0.831 & 0.799 & 0.807 & 0.803 & 0.780 & 0.792 & 0.786 & 0.803 & 0.746 & 0.774 \\
5 & 0.692 & 0.655 & 0.673 & 0.752 & 0.649 & 0.696 & 0.690 & 0.595 & 0.639 & 0.695 & 0.530 & 0.601 & 0.662 & 0.536 & 0.592 \\
6 & 0.771 & 0.698 & 0.733 & 0.767 & 0.704 & 0.734 & 0.769 & 0.711 & 0.739 & 0.747 & 0.704 & 0.725 & 0.719 & 0.755 & 0.736 \\
\hline
\end{tabular}
\end{sideways}
\end{table}
\clearpage

\subsection*{Congressional Bills - Issue Topic Classification (-,+,+)}
\begin{table}[H]
\centering
\setlength{\tabcolsep}{3pt}
\small
\caption{Congressional Bills - Issue Topic Classification Performance Metrics (GPT 4o) - (-,+,+)}
\begin{sideways}
\begin{tabular}{|l|c|c|c|c|c|c|c|c|c|c|c|c|c|c|c|}
\hline
Class & Precision & Recall & F1 & Precision & Recall & F1 & Precision & Recall & F1 & Precision & Recall & F1 & Precision & Recall & F1 \\
 & \multicolumn{3}{c|}{1} & \multicolumn{3}{c|}{10} & \multicolumn{3}{c|}{100} & \multicolumn{3}{c|}{500} & \multicolumn{3}{c|}{1000} \\
\hline
1 & 0.619 & 0.306 & 0.410 & 0.613 & 0.372 & 0.463 & 0.595 & 0.255 & 0.357 & 0.700 & 0.179 & 0.285 & 0.738 & 0.158 & 0.261 \\
2 & 0.681 & 0.660 & 0.670 & 0.690 & 0.617 & 0.652 & 0.797 & 0.585 & 0.675 & 0.797 & 0.500 & 0.614 & 0.875 & 0.447 & 0.592 \\
3 & 0.774 & 0.814 & 0.793 & 0.777 & 0.803 & 0.790 & 0.781 & 0.810 & 0.795 & 0.806 & 0.790 & 0.798 & 0.795 & 0.790 & 0.793 \\
4 & 0.819 & 0.756 & 0.786 & 0.832 & 0.756 & 0.793 & 0.814 & 0.777 & 0.795 & 0.798 & 0.741 & 0.768 & 0.801 & 0.695 & 0.745 \\
5 & 0.719 & 0.518 & 0.602 & 0.723 & 0.560 & 0.631 & 0.719 & 0.518 & 0.602 & 0.674 & 0.554 & 0.608 & 0.672 & 0.524 & 0.589 \\
6 & 0.758 & 0.711 & 0.734 & 0.745 & 0.717 & 0.731 & 0.756 & 0.742 & 0.749 & 0.737 & 0.704 & 0.720 & 0.714 & 0.723 & 0.719 \\
\hline
\end{tabular}
\end{sideways}
\end{table}
\clearpage

\subsection*{Congressional Bills - Issue Topic Classification (+,-,+)}
\begin{table}[H]
\centering
\setlength{\tabcolsep}{3pt}
\small
\caption{Congressional Bills - Issue Topic Classification Performance Metrics (GPT 4o) - (+,-,+)}
\begin{sideways}
\begin{tabular}{|l|c|c|c|c|c|c|c|c|c|c|c|c|c|c|c|}
\hline
Class & Precision & Recall & F1 & Precision & Recall & F1 & Precision & Recall & F1 & Precision & Recall & F1 & Precision & Recall & F1 \\
 & \multicolumn{3}{c|}{1} & \multicolumn{3}{c|}{10} & \multicolumn{3}{c|}{100} & \multicolumn{3}{c|}{500} & \multicolumn{3}{c|}{1000} \\
\hline
1 & 0.571 & 0.449 & 0.503 & 0.549 & 0.429 & 0.481 & 0.486 & 0.429 & 0.455 & 0.487 & 0.474 & 0.481 & 0.439 & 0.821 & 0.572 \\
2 & 0.670 & 0.649 & 0.659 & 0.726 & 0.649 & 0.685 & 0.742 & 0.521 & 0.613 & 0.782 & 0.457 & 0.577 & 0.848 & 0.415 & 0.557 \\
3 & 0.757 & 0.878 & 0.813 & 0.789 & 0.861 & 0.823 & 0.781 & 0.871 & 0.824 & 0.773 & 0.844 & 0.807 & 0.745 & 0.824 & 0.783 \\
4 & 0.811 & 0.827 & 0.819 & 0.836 & 0.827 & 0.832 & 0.821 & 0.792 & 0.806 & 0.804 & 0.772 & 0.788 & 0.825 & 0.741 & 0.781 \\
5 & 0.713 & 0.679 & 0.695 & 0.754 & 0.637 & 0.690 & 0.712 & 0.589 & 0.645 & 0.691 & 0.571 & 0.625 & 0.650 & 0.542 & 0.591 \\
6 & 0.777 & 0.811 & 0.794 & 0.767 & 0.786 & 0.776 & 0.753 & 0.767 & 0.760 & 0.728 & 0.774 & 0.750 & 0.707 & 0.742 & 0.724 \\
\hline
\end{tabular}
\end{sideways}
\end{table}
\clearpage

\subsection*{Congressional Bills - Issue Topic Classification (+,+,-)}
\begin{table}[H]
\centering
\setlength{\tabcolsep}{3pt}
\small
\caption{Congressional Bills - Issue Topic Classification Performance Metrics (GPT 4o) - (+,+,-)}
\begin{sideways}
\begin{tabular}{|l|c|c|c|c|c|c|c|c|c|c|c|c|c|c|c|}
\hline
Class & Precision & Recall & F1 & Precision & Recall & F1 & Precision & Recall & F1 & Precision & Recall & F1 & Precision & Recall & F1 \\
 & \multicolumn{3}{c|}{1} & \multicolumn{3}{c|}{10} & \multicolumn{3}{c|}{100} & \multicolumn{3}{c|}{500} & \multicolumn{3}{c|}{1000} \\
\hline
1 & 0.664 & 0.444 & 0.532 & 0.664 & 0.413 & 0.509 & 0.446 & 0.418 & 0.432 & 0.449 & 0.566 & 0.501 & 0.434 & 0.750 & 0.550 \\
2 & 0.836 & 0.649 & 0.731 & 0.789 & 0.638 & 0.706 & 0.831 & 0.574 & 0.679 & 0.781 & 0.532 & 0.633 & 0.885 & 0.489 & 0.630 \\
3 & 0.825 & 0.814 & 0.819 & 0.814 & 0.800 & 0.807 & 0.810 & 0.810 & 0.810 & 0.800 & 0.814 & 0.807 & 0.810 & 0.780 & 0.794 \\
4 & 0.861 & 0.756 & 0.805 & 0.833 & 0.736 & 0.782 & 0.828 & 0.756 & 0.790 & 0.805 & 0.711 & 0.755 & 0.815 & 0.695 & 0.751 \\
5 & 0.816 & 0.554 & 0.660 & 0.821 & 0.601 & 0.694 & 0.790 & 0.560 & 0.655 & 0.693 & 0.524 & 0.597 & 0.714 & 0.536 & 0.612 \\
6 & 0.841 & 0.698 & 0.763 & 0.812 & 0.654 & 0.725 & 0.768 & 0.667 & 0.714 & 0.772 & 0.704 & 0.737 & 0.778 & 0.704 & 0.739 \\
\hline
\end{tabular}
\end{sideways}
\end{table}
\clearpage

\subsection*{Congressional Bills - Issue Topic Classification (+,+,+)}
\begin{table}[H]
\centering
\setlength{\tabcolsep}{3pt}
\small
\caption{Congressional Bills - Issue Topic Classification Performance Metrics (GPT 4o) - (+,+,+)}
\begin{sideways}
\begin{tabular}{|l|c|c|c|c|c|c|c|c|c|c|c|c|c|c|c|}
\hline
Class & Precision & Recall & F1 & Precision & Recall & F1 & Precision & Recall & F1 & Precision & Recall & F1 & Precision & Recall & F1 \\
 & \multicolumn{3}{c|}{1} & \multicolumn{3}{c|}{10} & \multicolumn{3}{c|}{100} & \multicolumn{3}{c|}{500} & \multicolumn{3}{c|}{1000} \\
\hline
1 & 0.580 & 0.423 & 0.490 & 0.566 & 0.413 & 0.478 & 0.530 & 0.316 & 0.396 & 0.525 & 0.434 & 0.475 & 0.465 & 0.684 & 0.554 \\
2 & 0.721 & 0.660 & 0.689 & 0.739 & 0.691 & 0.714 & 0.773 & 0.617 & 0.686 & 0.841 & 0.564 & 0.675 & 0.845 & 0.521 & 0.645 \\
3 & 0.768 & 0.786 & 0.777 & 0.791 & 0.797 & 0.794 & 0.803 & 0.817 & 0.810 & 0.803 & 0.786 & 0.795 & 0.804 & 0.807 & 0.805 \\
4 & 0.841 & 0.777 & 0.807 & 0.827 & 0.777 & 0.801 & 0.839 & 0.766 & 0.801 & 0.814 & 0.756 & 0.784 & 0.807 & 0.741 & 0.772 \\
5 & 0.797 & 0.583 & 0.674 & 0.810 & 0.607 & 0.694 & 0.754 & 0.565 & 0.646 & 0.717 & 0.542 & 0.617 & 0.711 & 0.542 & 0.615 \\
6 & 0.806 & 0.704 & 0.752 & 0.789 & 0.704 & 0.744 & 0.786 & 0.717 & 0.750 & 0.753 & 0.711 & 0.731 & 0.722 & 0.736 & 0.729 \\
\hline
\end{tabular}
\end{sideways}
\end{table}
\clearpage

\subsection*{Congressional Bills - Issue Topic Classification (-,-,-)}
\begin{table}[H]
\centering
\setlength{\tabcolsep}{3pt}
\small
\caption{Congressional Bills - Issue Topic Classification Performance Metrics (Gemini 2.0 Flash) - (-,-,-)}
\begin{sideways}
\begin{tabular}{|l|c|c|c|c|c|c|c|c|c|c|c|c|c|c|c|}
\hline
Class & Precision & Recall & F1 & Precision & Recall & F1 & Precision & Recall & F1 & Precision & Recall & F1 & Precision & Recall & F1 \\
 & \multicolumn{3}{c|}{1} & \multicolumn{3}{c|}{10} & \multicolumn{3}{c|}{100} & \multicolumn{3}{c|}{500} & \multicolumn{3}{c|}{1000} \\
\hline
1 & 0.533 & 0.617 & 0.572 & 0.578 & 0.587 & 0.582 & 0.505 & 0.719 & 0.594 & 0.435 & 0.735 & 0.546 & 0.448 & 0.770 & 0.567 \\
2 & 0.682 & 0.479 & 0.562 & 0.690 & 0.426 & 0.526 & 0.763 & 0.479 & 0.588 & 0.787 & 0.511 & 0.619 & 0.803 & 0.521 & 0.632 \\
3 & 0.760 & 0.878 & 0.814 & 0.738 & 0.861 & 0.795 & 0.764 & 0.854 & 0.806 & 0.768 & 0.831 & 0.798 & 0.742 & 0.800 & 0.770 \\
4 & 0.659 & 0.843 & 0.739 & 0.694 & 0.817 & 0.751 & 0.718 & 0.777 & 0.746 & 0.742 & 0.746 & 0.744 & 0.763 & 0.736 & 0.749 \\
5 & 0.406 & 0.744 & 0.525 & 0.417 & 0.798 & 0.548 & 0.463 & 0.708 & 0.560 & 0.566 & 0.667 & 0.612 & 0.618 & 0.637 & 0.628 \\
6 & 0.746 & 0.811 & 0.777 & 0.707 & 0.818 & 0.758 & 0.728 & 0.824 & 0.773 & 0.753 & 0.805 & 0.778 & 0.762 & 0.824 & 0.792 \\
\hline
\end{tabular}
\end{sideways}
\end{table}
\clearpage

\subsection*{Congressional Bills - Issue Topic Classification (-,-,+)}
\begin{table}[H]
\centering
\setlength{\tabcolsep}{3pt}
\small
\caption{Congressional Bills - Issue Topic Classification Performance Metrics (Gemini 2.0 Flash) - (-,-,+)}
\begin{sideways}
\begin{tabular}{|l|c|c|c|c|c|c|c|c|c|c|c|c|c|c|c|}
\hline
Class & Precision & Recall & F1 & Precision & Recall & F1 & Precision & Recall & F1 & Precision & Recall & F1 & Precision & Recall & F1 \\
 & \multicolumn{3}{c|}{1} & \multicolumn{3}{c|}{10} & \multicolumn{3}{c|}{100} & \multicolumn{3}{c|}{500} & \multicolumn{3}{c|}{1000} \\
\hline
1 & 0.703 & 0.327 & 0.446 & 0.702 & 0.337 & 0.455 & 0.600 & 0.367 & 0.456 & 0.548 & 0.408 & 0.468 & 0.587 & 0.413 & 0.485 \\
2 & 0.710 & 0.521 & 0.601 & 0.721 & 0.521 & 0.605 & 0.800 & 0.553 & 0.654 & 0.732 & 0.553 & 0.630 & 0.800 & 0.511 & 0.623 \\
3 & 0.783 & 0.834 & 0.808 & 0.773 & 0.844 & 0.807 & 0.788 & 0.831 & 0.809 & 0.779 & 0.875 & 0.824 & 0.761 & 0.810 & 0.785 \\
4 & 0.717 & 0.822 & 0.766 & 0.734 & 0.812 & 0.771 & 0.800 & 0.792 & 0.796 & 0.799 & 0.746 & 0.772 & 0.763 & 0.766 & 0.765 \\
5 & 0.467 & 0.762 & 0.579 & 0.488 & 0.726 & 0.584 & 0.529 & 0.643 & 0.581 & 0.639 & 0.601 & 0.620 & 0.625 & 0.536 & 0.577 \\
6 & 0.757 & 0.805 & 0.780 & 0.740 & 0.824 & 0.780 & 0.739 & 0.818 & 0.776 & 0.743 & 0.818 & 0.778 & 0.740 & 0.805 & 0.771 \\
\hline
\end{tabular}
\end{sideways}
\end{table}
\clearpage

\subsection*{Congressional Bills - Issue Topic Classification (-,+,-)}
\begin{table}[H]
\centering
\setlength{\tabcolsep}{3pt}
\small
\caption{Congressional Bills - Issue Topic Classification Performance Metrics (Gemini 2.0 Flash) - (-,+,-)}
\begin{sideways}
\begin{tabular}{|l|c|c|c|c|c|c|c|c|c|c|c|c|c|c|c|}
\hline
Class & Precision & Recall & F1 & Precision & Recall & F1 & Precision & Recall & F1 & Precision & Recall & F1 & Precision & Recall & F1 \\
 & \multicolumn{3}{c|}{1} & \multicolumn{3}{c|}{10} & \multicolumn{3}{c|}{100} & \multicolumn{3}{c|}{500} & \multicolumn{3}{c|}{1000} \\
\hline
1 & 0.772 & 0.224 & 0.348 & 0.758 & 0.255 & 0.382 & 0.638 & 0.189 & 0.291 & 0.525 & 0.270 & 0.357 & 0.492 & 0.444 & 0.466 \\
2 & 0.817 & 0.521 & 0.636 & 0.770 & 0.500 & 0.606 & 0.852 & 0.553 & 0.671 & 0.836 & 0.543 & 0.658 & 0.841 & 0.564 & 0.675 \\
3 & 0.822 & 0.797 & 0.809 & 0.794 & 0.759 & 0.776 & 0.803 & 0.732 & 0.766 & 0.812 & 0.749 & 0.780 & 0.787 & 0.739 & 0.762 \\
4 & 0.700 & 0.792 & 0.743 & 0.744 & 0.782 & 0.762 & 0.768 & 0.756 & 0.762 & 0.757 & 0.711 & 0.733 & 0.769 & 0.711 & 0.739 \\
5 & 0.575 & 0.756 & 0.653 & 0.571 & 0.744 & 0.646 & 0.646 & 0.696 & 0.670 & 0.699 & 0.649 & 0.673 & 0.698 & 0.673 & 0.685 \\
6 & 0.796 & 0.686 & 0.736 & 0.769 & 0.711 & 0.739 & 0.775 & 0.692 & 0.731 & 0.801 & 0.686 & 0.739 & 0.793 & 0.698 & 0.742 \\
\hline
\end{tabular}
\end{sideways}
\end{table}
\clearpage

\subsection*{Congressional Bills - Issue Topic Classification (+,-,-)}
\begin{table}[H]
\centering
\setlength{\tabcolsep}{3pt}
\small
\caption{Congressional Bills - Issue Topic Classification Performance Metrics (Gemini 2.0 Flash) - (+,-,-)}
\begin{sideways}
\begin{tabular}{|l|c|c|c|c|c|c|c|c|c|c|c|c|c|c|c|}
\hline
Class & Precision & Recall & F1 & Precision & Recall & F1 & Precision & Recall & F1 & Precision & Recall & F1 & Precision & Recall & F1 \\
 & \multicolumn{3}{c|}{1} & \multicolumn{3}{c|}{10} & \multicolumn{3}{c|}{100} & \multicolumn{3}{c|}{500} & \multicolumn{3}{c|}{1000} \\
\hline
1 & 0.435 & 0.816 & 0.567 & 0.449 & 0.816 & 0.580 & 0.422 & 0.837 & 0.561 & 0.416 & 0.862 & 0.561 & 0.427 & 0.704 & 0.532 \\
2 & 0.814 & 0.606 & 0.695 & 0.875 & 0.596 & 0.709 & 0.845 & 0.638 & 0.727 & 0.810 & 0.543 & 0.650 & 0.865 & 0.479 & 0.616 \\
3 & 0.782 & 0.864 & 0.821 & 0.805 & 0.800 & 0.803 & 0.816 & 0.841 & 0.828 & 0.798 & 0.817 & 0.807 & 0.796 & 0.780 & 0.788 \\
4 & 0.776 & 0.827 & 0.801 & 0.759 & 0.817 & 0.787 & 0.776 & 0.843 & 0.808 & 0.754 & 0.777 & 0.765 & 0.793 & 0.721 & 0.755 \\
5 & 0.464 & 0.720 & 0.564 & 0.487 & 0.679 & 0.567 & 0.536 & 0.661 & 0.592 & 0.641 & 0.637 & 0.639 & 0.692 & 0.548 & 0.611 \\
6 & 0.790 & 0.616 & 0.693 & 0.793 & 0.698 & 0.742 & 0.790 & 0.711 & 0.748 & 0.784 & 0.660 & 0.717 & 0.765 & 0.635 & 0.694 \\
\hline
\end{tabular}
\end{sideways}
\end{table}
\clearpage

\subsection*{Congressional Bills - Issue Topic Classification (-,+,+)}
\begin{table}[H]
\centering
\setlength{\tabcolsep}{3pt}
\small
\caption{Congressional Bills - Issue Topic Classification Performance Metrics (Gemini 2.0 Flash) - (-,+,+)}
\begin{sideways}
\begin{tabular}{|l|c|c|c|c|c|c|c|c|c|c|c|c|c|c|c|}
\hline
Class & Precision & Recall & F1 & Precision & Recall & F1 & Precision & Recall & F1 & Precision & Recall & F1 & Precision & Recall & F1 \\
 & \multicolumn{3}{c|}{1} & \multicolumn{3}{c|}{10} & \multicolumn{3}{c|}{100} & \multicolumn{3}{c|}{500} & \multicolumn{3}{c|}{1000} \\
\hline
1 & 0.750 & 0.184 & 0.295 & 0.734 & 0.240 & 0.362 & 0.671 & 0.270 & 0.385 & 0.597 & 0.235 & 0.337 & 0.593 & 0.357 & 0.446 \\
2 & 0.771 & 0.574 & 0.659 & 0.750 & 0.638 & 0.690 & 0.795 & 0.617 & 0.695 & 0.768 & 0.564 & 0.650 & 0.821 & 0.585 & 0.683 \\
3 & 0.838 & 0.739 & 0.786 & 0.826 & 0.742 & 0.782 & 0.811 & 0.742 & 0.775 & 0.806 & 0.759 & 0.782 & 0.801 & 0.766 & 0.783 \\
4 & 0.802 & 0.782 & 0.792 & 0.800 & 0.772 & 0.786 & 0.808 & 0.726 & 0.765 & 0.793 & 0.721 & 0.755 & 0.807 & 0.741 & 0.772 \\
5 & 0.665 & 0.625 & 0.644 & 0.653 & 0.649 & 0.651 & 0.675 & 0.631 & 0.652 & 0.716 & 0.571 & 0.636 & 0.711 & 0.542 & 0.615 \\
6 & 0.812 & 0.679 & 0.740 & 0.786 & 0.692 & 0.736 & 0.790 & 0.711 & 0.748 & 0.778 & 0.704 & 0.739 & 0.799 & 0.723 & 0.759 \\
\hline
\end{tabular}
\end{sideways}
\end{table}
\clearpage

\subsection*{Congressional Bills - Issue Topic Classification (+,-,+)}
\begin{table}[H]
\centering
\setlength{\tabcolsep}{3pt}
\small
\caption{Congressional Bills - Issue Topic Classification Performance Metrics (Gemini 2.0 Flash) - (+,-,+)}
\begin{sideways}
\begin{tabular}{|l|c|c|c|c|c|c|c|c|c|c|c|c|c|c|c|}
\hline
Class & Precision & Recall & F1 & Precision & Recall & F1 & Precision & Recall & F1 & Precision & Recall & F1 & Precision & Recall & F1 \\
 & \multicolumn{3}{c|}{1} & \multicolumn{3}{c|}{10} & \multicolumn{3}{c|}{100} & \multicolumn{3}{c|}{500} & \multicolumn{3}{c|}{1000} \\
\hline
1 & 0.533 & 0.571 & 0.552 & 0.577 & 0.571 & 0.574 & 0.498 & 0.551 & 0.523 & 0.493 & 0.510 & 0.501 & 0.513 & 0.520 & 0.516 \\
2 & 0.794 & 0.574 & 0.667 & 0.779 & 0.638 & 0.702 & 0.811 & 0.638 & 0.714 & 0.833 & 0.585 & 0.688 & 0.815 & 0.564 & 0.667 \\
3 & 0.796 & 0.844 & 0.819 & 0.807 & 0.820 & 0.813 & 0.815 & 0.837 & 0.826 & 0.792 & 0.864 & 0.827 & 0.759 & 0.888 & 0.819 \\
4 & 0.788 & 0.832 & 0.810 & 0.796 & 0.812 & 0.804 & 0.806 & 0.822 & 0.814 & 0.801 & 0.797 & 0.799 & 0.800 & 0.772 & 0.786 \\
5 & 0.569 & 0.732 & 0.641 & 0.546 & 0.738 & 0.628 & 0.557 & 0.637 & 0.594 & 0.635 & 0.631 & 0.633 & 0.654 & 0.631 & 0.642 \\
6 & 0.820 & 0.686 & 0.747 & 0.799 & 0.723 & 0.759 & 0.788 & 0.774 & 0.781 & 0.769 & 0.755 & 0.762 & 0.767 & 0.767 & 0.767 \\
\hline
\end{tabular}
\end{sideways}
\end{table}
\clearpage

\subsection*{Congressional Bills - Issue Topic Classification (+,+,-)}
\begin{table}[H]
\centering
\setlength{\tabcolsep}{3pt}
\small
\caption{Congressional Bills - Issue Topic Classification Performance Metrics (Gemini 2.0 Flash) - (+,+,-)}
\begin{sideways}
\begin{tabular}{|l|c|c|c|c|c|c|c|c|c|c|c|c|c|c|c|}
\hline
Class & Precision & Recall & F1 & Precision & Recall & F1 & Precision & Recall & F1 & Precision & Recall & F1 & Precision & Recall & F1 \\
 & \multicolumn{3}{c|}{1} & \multicolumn{3}{c|}{10} & \multicolumn{3}{c|}{100} & \multicolumn{3}{c|}{500} & \multicolumn{3}{c|}{1000} \\
\hline
1 & 0.586 & 0.607 & 0.596 & 0.527 & 0.699 & 0.601 & 0.454 & 0.658 & 0.537 & 0.467 & 0.694 & 0.559 & 0.465 & 0.811 & 0.591 \\
2 & 0.869 & 0.564 & 0.684 & 0.836 & 0.596 & 0.696 & 0.906 & 0.617 & 0.734 & 0.815 & 0.564 & 0.667 & 0.836 & 0.543 & 0.658 \\
3 & 0.815 & 0.820 & 0.818 & 0.822 & 0.783 & 0.802 & 0.844 & 0.786 & 0.814 & 0.816 & 0.810 & 0.813 & 0.810 & 0.810 & 0.810 \\
4 & 0.758 & 0.746 & 0.752 & 0.804 & 0.812 & 0.808 & 0.800 & 0.812 & 0.806 & 0.788 & 0.756 & 0.772 & 0.791 & 0.751 & 0.771 \\
5 & 0.643 & 0.762 & 0.698 & 0.642 & 0.726 & 0.682 & 0.728 & 0.685 & 0.706 & 0.772 & 0.685 & 0.726 & 0.779 & 0.649 & 0.708 \\
6 & 0.843 & 0.673 & 0.748 & 0.843 & 0.642 & 0.729 & 0.827 & 0.660 & 0.734 & 0.800 & 0.629 & 0.704 & 0.795 & 0.635 & 0.706 \\
\hline
\end{tabular}
\end{sideways}
\end{table}
\clearpage

\subsection*{Congressional Bills - Issue Topic Classification (+,+,+)}
\begin{table}[H]
\centering
\setlength{\tabcolsep}{3pt}
\small
\caption{Congressional Bills - Issue Topic Classification Performance Metrics (Gemini 2.0 Flash) - (+,+,+)}
\begin{sideways}
\begin{tabular}{|l|c|c|c|c|c|c|c|c|c|c|c|c|c|c|c|}
\hline
Class & Precision & Recall & F1 & Precision & Recall & F1 & Precision & Recall & F1 & Precision & Recall & F1 & Precision & Recall & F1 \\
 & \multicolumn{3}{c|}{1} & \multicolumn{3}{c|}{10} & \multicolumn{3}{c|}{100} & \multicolumn{3}{c|}{500} & \multicolumn{3}{c|}{1000} \\
\hline
1 & 0.480 & 0.480 & 0.480 & 0.607 & 0.520 & 0.560 & 0.547 & 0.531 & 0.539 & 0.534 & 0.556 & 0.545 & 0.523 & 0.526 & 0.524 \\
2 & 0.845 & 0.638 & 0.727 & 0.816 & 0.660 & 0.729 & 0.847 & 0.649 & 0.735 & 0.848 & 0.596 & 0.700 & 0.841 & 0.564 & 0.675 \\
3 & 0.829 & 0.790 & 0.809 & 0.836 & 0.759 & 0.796 & 0.835 & 0.773 & 0.803 & 0.829 & 0.790 & 0.809 & 0.812 & 0.807 & 0.810 \\
4 & 0.815 & 0.782 & 0.798 & 0.828 & 0.782 & 0.804 & 0.817 & 0.792 & 0.804 & 0.807 & 0.766 & 0.786 & 0.839 & 0.766 & 0.801 \\
5 & 0.665 & 0.685 & 0.674 & 0.670 & 0.702 & 0.686 & 0.709 & 0.637 & 0.671 & 0.754 & 0.619 & 0.680 & 0.752 & 0.595 & 0.664 \\
6 & 0.858 & 0.648 & 0.738 & 0.837 & 0.679 & 0.750 & 0.828 & 0.698 & 0.758 & 0.784 & 0.686 & 0.732 & 0.803 & 0.692 & 0.743 \\
\hline
\end{tabular}
\end{sideways}
\end{table}
\clearpage

\subsection*{ANES (MIP) - Emotionality Classification (-,-,-)}
\setlength{\tabcolsep}{3pt}
\begin{longtable}{|l|c|c|c|c|c|c|c|c|c|c|c|c|}
\caption{ANES (MIP) - Emotionality Classification Performance Metrics (Gemini 2.0 Flash) - (-,-,-)}\\
\toprule
Class & Precision & Recall & F1 & Precision & Recall & F1 & Precision & Recall & F1 & Precision & Recall & F1 \\
 & \multicolumn{3}{c|}{1} & \multicolumn{3}{c|}{10} & \multicolumn{3}{c|}{100} & \multicolumn{3}{c|}{300} \\
\midrule
\endfirsthead
\caption[]{ANES (MIP) - Emotionality Classification Performance Metrics (Gemini 2.0 Flash) - (-,-,-) (continued)}\\
\toprule
Class & Precision & Recall & F1 & Precision & Recall & F1 & Precision & Recall & F1 & Precision & Recall & F1 \\
 & \multicolumn{3}{c|}{1} & \multicolumn{3}{c|}{10} & \multicolumn{3}{c|}{100} & \multicolumn{3}{c|}{300} \\
\midrule
\endhead
\midrule
\multicolumn{13}{r}{\textit{Continued on next page}}\\
\midrule
\endfoot
\bottomrule
\endlastfoot

0 & 0.963 & 0.339 & 0.502 & 0.969 & 0.692 & 0.808 & 0.957 & 0.869 & 0.911 & 0.960 & 0.875 & 0.915 \\
1 & 0.310 & 0.958 & 0.469 & 0.484 & 0.930 & 0.636 & 0.674 & 0.873 & 0.761 & 0.688 & 0.881 & 0.772 \\
\end{longtable}

\subsection*{ANES (MIP) - Emotionality Classification (-,-,+)}
\setlength{\tabcolsep}{3pt}
\begin{longtable}{|l|c|c|c|c|c|c|c|c|c|c|c|c|}
\caption{ANES (MIP) - Emotionality Classification Performance Metrics (Gemini 2.0 Flash) - (-,-,+)}\\
\toprule
Class & Precision & Recall & F1 & Precision & Recall & F1 & Precision & Recall & F1 & Precision & Recall & F1 \\
 & \multicolumn{3}{c|}{1} & \multicolumn{3}{c|}{10} & \multicolumn{3}{c|}{100} & \multicolumn{3}{c|}{300} \\
\midrule
\endfirsthead
\caption[]{ANES (MIP) - Emotionality Classification Performance Metrics (Gemini 2.0 Flash) - (-,-,+) (continued)}\\
\toprule
Class & Precision & Recall & F1 & Precision & Recall & F1 & Precision & Recall & F1 & Precision & Recall & F1 \\
 & \multicolumn{3}{c|}{1} & \multicolumn{3}{c|}{10} & \multicolumn{3}{c|}{100} & \multicolumn{3}{c|}{300} \\
\midrule
\endhead
\midrule
\multicolumn{13}{r}{\textit{Continued on next page}}\\
\midrule
\endfoot
\bottomrule
\endlastfoot

0 & 0.981 & 0.688 & 0.809 & 0.954 & 0.824 & 0.885 & 0.966 & 0.856 & 0.907 & 0.957 & 0.885 & 0.920 \\
1 & 0.488 & 0.958 & 0.646 & 0.607 & 0.873 & 0.716 & 0.659 & 0.901 & 0.762 & 0.702 & 0.871 & 0.777 \\
\end{longtable}

\subsection*{ANES (MIP) - Emotionality Classification (-,+,-)}
\setlength{\tabcolsep}{3pt}
\begin{longtable}{|l|c|c|c|c|c|c|c|c|c|c|c|c|}
\caption{ANES (MIP) - Emotionality Classification Performance Metrics (Gemini 2.0 Flash) - (-,+,-)}\\
\toprule
Class & Precision & Recall & F1 & Precision & Recall & F1 & Precision & Recall & F1 & Precision & Recall & F1 \\
 & \multicolumn{3}{c|}{1} & \multicolumn{3}{c|}{10} & \multicolumn{3}{c|}{100} & \multicolumn{3}{c|}{300} \\
\midrule
\endfirsthead
\caption[]{ANES (MIP) - Emotionality Classification Performance Metrics (Gemini 2.0 Flash) - (-,+,-) (continued)}\\
\toprule
Class & Precision & Recall & F1 & Precision & Recall & F1 & Precision & Recall & F1 & Precision & Recall & F1 \\
 & \multicolumn{3}{c|}{1} & \multicolumn{3}{c|}{10} & \multicolumn{3}{c|}{100} & \multicolumn{3}{c|}{300} \\
\midrule
\endhead
\midrule
\multicolumn{13}{r}{\textit{Continued on next page}}\\
\midrule
\endfoot
\bottomrule
\endlastfoot

0 & 0.977 & 0.551 & 0.705 & 0.963 & 0.730 & 0.830 & 0.959 & 0.872 & 0.914 & 0.960 & 0.891 & 0.924 \\
1 & 0.398 & 0.958 & 0.562 & 0.511 & 0.909 & 0.654 & 0.682 & 0.880 & 0.768 & 0.715 & 0.880 & 0.789 \\
\end{longtable}

\subsection*{ANES (MIP) - Emotionality Classification (+,-,-)}
\setlength{\tabcolsep}{3pt}
\begin{longtable}{|l|c|c|c|c|c|c|c|c|c|c|c|c|}
\caption{ANES (MIP) - Emotionality Classification Performance Metrics (Gemini 2.0 Flash) - (+,-,-)}\\
\toprule
Class & Precision & Recall & F1 & Precision & Recall & F1 & Precision & Recall & F1 & Precision & Recall & F1 \\
 & \multicolumn{3}{c|}{1} & \multicolumn{3}{c|}{10} & \multicolumn{3}{c|}{100} & \multicolumn{3}{c|}{300} \\
\midrule
\endfirsthead
\caption[]{ANES (MIP) - Emotionality Classification Performance Metrics (Gemini 2.0 Flash) - (+,-,-) (continued)}\\
\toprule
Class & Precision & Recall & F1 & Precision & Recall & F1 & Precision & Recall & F1 & Precision & Recall & F1 \\
 & \multicolumn{3}{c|}{1} & \multicolumn{3}{c|}{10} & \multicolumn{3}{c|}{100} & \multicolumn{3}{c|}{300} \\
\midrule
\endhead
\midrule
\multicolumn{13}{r}{\textit{Continued on next page}}\\
\midrule
\endfoot
\bottomrule
\endlastfoot

0 & 0.973 & 0.639 & 0.772 & 0.944 & 0.853 & 0.896 & 0.883 & 0.952 & 0.916 & 0.873 & 0.968 & 0.918 \\
1 & 0.448 & 0.944 & 0.608 & 0.638 & 0.837 & 0.724 & 0.793 & 0.594 & 0.679 & 0.842 & 0.545 & 0.661 \\
\end{longtable}

\subsection*{ANES (MIP) - Emotionality Classification (-,+,+)}
\setlength{\tabcolsep}{3pt}
\begin{longtable}{|l|c|c|c|c|c|c|c|c|c|c|c|c|}
\caption{ANES (MIP) - Emotionality Classification Performance Metrics (Gemini 2.0 Flash) - (-,+,+)}\\
\toprule
Class & Precision & Recall & F1 & Precision & Recall & F1 & Precision & Recall & F1 & Precision & Recall & F1 \\
 & \multicolumn{3}{c|}{1} & \multicolumn{3}{c|}{10} & \multicolumn{3}{c|}{100} & \multicolumn{3}{c|}{300} \\
\midrule
\endfirsthead
\caption[]{ANES (MIP) - Emotionality Classification Performance Metrics (Gemini 2.0 Flash) - (-,+,+) (continued)}\\
\toprule
Class & Precision & Recall & F1 & Precision & Recall & F1 & Precision & Recall & F1 & Precision & Recall & F1 \\
 & \multicolumn{3}{c|}{1} & \multicolumn{3}{c|}{10} & \multicolumn{3}{c|}{100} & \multicolumn{3}{c|}{300} \\
\midrule
\endhead
\midrule
\multicolumn{13}{r}{\textit{Continued on next page}}\\
\midrule
\endfoot
\bottomrule
\endlastfoot

0 & 0.969 & 0.675 & 0.796 & 0.945 & 0.874 & 0.908 & 0.943 & 0.864 & 0.902 & 0.959 & 0.850 & 0.901 \\
1 & 0.470 & 0.930 & 0.624 & 0.674 & 0.837 & 0.746 & 0.655 & 0.831 & 0.733 & 0.646 & 0.881 & 0.746 \\
\end{longtable}

\subsection*{ANES (MIP) - Emotionality Classification (+,-,+)}
\setlength{\tabcolsep}{3pt}
\begin{longtable}{|l|c|c|c|c|c|c|c|c|c|c|c|c|}
\caption{ANES (MIP) - Emotionality Classification Performance Metrics (Gemini 2.0 Flash) - (+,-,+)}\\
\toprule
Class & Precision & Recall & F1 & Precision & Recall & F1 & Precision & Recall & F1 & Precision & Recall & F1 \\
 & \multicolumn{3}{c|}{1} & \multicolumn{3}{c|}{10} & \multicolumn{3}{c|}{100} & \multicolumn{3}{c|}{300} \\
\midrule
\endfirsthead
\caption[]{ANES (MIP) - Emotionality Classification Performance Metrics (Gemini 2.0 Flash) - (+,-,+) (continued)}\\
\toprule
Class & Precision & Recall & F1 & Precision & Recall & F1 & Precision & Recall & F1 & Precision & Recall & F1 \\
 & \multicolumn{3}{c|}{1} & \multicolumn{3}{c|}{10} & \multicolumn{3}{c|}{100} & \multicolumn{3}{c|}{300} \\
\midrule
\endhead
\midrule
\multicolumn{13}{r}{\textit{Continued on next page}}\\
\midrule
\endfoot
\bottomrule
\endlastfoot

0 & 0.973 & 0.792 & 0.873 & 0.948 & 0.893 & 0.919 & 0.941 & 0.908 & 0.924 & 0.875 & 0.974 & 0.922 \\
1 & 0.581 & 0.930 & 0.715 & 0.709 & 0.842 & 0.770 & 0.734 & 0.817 & 0.773 & 0.867 & 0.552 & 0.675 \\
\end{longtable}

\subsection*{ANES (MIP) - Emotionality Classification (+,+,-)}
\setlength{\tabcolsep}{3pt}
\begin{longtable}{|l|c|c|c|c|c|c|c|c|c|c|c|c|}
\caption{ANES (MIP) - Emotionality Classification Performance Metrics (Gemini 2.0 Flash) - (+,+,-)}\\
\toprule
Class & Precision & Recall & F1 & Precision & Recall & F1 & Precision & Recall & F1 & Precision & Recall & F1 \\
 & \multicolumn{3}{c|}{1} & \multicolumn{3}{c|}{10} & \multicolumn{3}{c|}{100} & \multicolumn{3}{c|}{300} \\
\midrule
\endfirsthead
\caption[]{ANES (MIP) - Emotionality Classification Performance Metrics (Gemini 2.0 Flash) - (+,+,-) (continued)}\\
\toprule
Class & Precision & Recall & F1 & Precision & Recall & F1 & Precision & Recall & F1 & Precision & Recall & F1 \\
 & \multicolumn{3}{c|}{1} & \multicolumn{3}{c|}{10} & \multicolumn{3}{c|}{100} & \multicolumn{3}{c|}{300} \\
\midrule
\endhead
\midrule
\multicolumn{13}{r}{\textit{Continued on next page}}\\
\midrule
\endfoot
\bottomrule
\endlastfoot

0 & 0.961 & 0.755 & 0.846 & 0.945 & 0.854 & 0.897 & 0.885 & 0.952 & 0.917 & 0.847 & 0.974 & 0.906 \\
1 & 0.533 & 0.901 & 0.670 & 0.641 & 0.839 & 0.727 & 0.795 & 0.600 & 0.684 & 0.843 & 0.431 & 0.566 \\
\end{longtable}

\subsection*{ANES (MIP) - Emotionality Classification (+,+,+)}
\setlength{\tabcolsep}{3pt}
\begin{longtable}{|l|c|c|c|c|c|c|c|c|c|c|c|c|}
\caption{ANES (MIP) - Emotionality Classification Performance Metrics (Gemini 2.0 Flash) - (+,+,+)}\\
\toprule
Class & Precision & Recall & F1 & Precision & Recall & F1 & Precision & Recall & F1 & Precision & Recall & F1 \\
 & \multicolumn{3}{c|}{1} & \multicolumn{3}{c|}{10} & \multicolumn{3}{c|}{100} & \multicolumn{3}{c|}{300} \\
\midrule
\endfirsthead
\caption[]{ANES (MIP) - Emotionality Classification Performance Metrics (Gemini 2.0 Flash) - (+,+,+) (continued)}\\
\toprule
Class & Precision & Recall & F1 & Precision & Recall & F1 & Precision & Recall & F1 & Precision & Recall & F1 \\
 & \multicolumn{3}{c|}{1} & \multicolumn{3}{c|}{10} & \multicolumn{3}{c|}{100} & \multicolumn{3}{c|}{300} \\
\midrule
\endhead
\midrule
\multicolumn{13}{r}{\textit{Continued on next page}}\\
\midrule
\endfoot
\bottomrule
\endlastfoot

0 & 0.951 & 0.900 & 0.925 & 0.948 & 0.930 & 0.939 & 0.925 & 0.941 & 0.933 & 0.851 & 0.981 & 0.911 \\
1 & 0.725 & 0.852 & 0.783 & 0.788 & 0.836 & 0.812 & 0.798 & 0.754 & 0.775 & 0.881 & 0.444 & 0.583 \\
\end{longtable}

\subsection*{ANES (MIP) - Emotionality Classification (-,-,-)}
\setlength{\tabcolsep}{3pt}
\begin{longtable}{|l|c|c|c|c|c|c|c|c|c|c|c|c|}
\caption{ANES (MIP) - Emotionality Classification Performance Metrics (GPT 4o) - (-,-,-)}\\
\toprule
Class & Precision & Recall & F1 & Precision & Recall & F1 & Precision & Recall & F1 & Precision & Recall & F1 \\
 & \multicolumn{3}{c|}{1} & \multicolumn{3}{c|}{10} & \multicolumn{3}{c|}{100} & \multicolumn{3}{c|}{300} \\
\midrule
\endfirsthead
\caption[]{ANES (MIP) - Emotionality Classification Performance Metrics (GPT 4o) - (-,-,-) (continued)}\\
\toprule
Class & Precision & Recall & F1 & Precision & Recall & F1 & Precision & Recall & F1 & Precision & Recall & F1 \\
 & \multicolumn{3}{c|}{1} & \multicolumn{3}{c|}{10} & \multicolumn{3}{c|}{100} & \multicolumn{3}{c|}{300} \\
\midrule
\endhead
\midrule
\multicolumn{13}{r}{\textit{Continued on next page}}\\
\midrule
\endfoot
\bottomrule
\endlastfoot

0 & 0.968 & 0.660 & 0.785 & 0.986 & 0.706 & 0.823 & 0.968 & 0.768 & 0.856 & 0.960 & 0.788 & 0.865 \\
1 & 0.459 & 0.930 & 0.615 & 0.505 & 0.967 & 0.663 & 0.551 & 0.918 & 0.689 & 0.570 & 0.894 & 0.695 \\
\end{longtable}

\subsection*{ANES (MIP) - Emotionality Classification (-,-,+)}
\setlength{\tabcolsep}{3pt}
\begin{longtable}{|l|c|c|c|c|c|c|c|c|c|c|c|c|}
\caption{ANES (MIP) - Emotionality Classification Performance Metrics (GPT 4o) - (-,-,+)}\\
\toprule
Class & Precision & Recall & F1 & Precision & Recall & F1 & Precision & Recall & F1 & Precision & Recall & F1 \\
 & \multicolumn{3}{c|}{1} & \multicolumn{3}{c|}{10} & \multicolumn{3}{c|}{100} & \multicolumn{3}{c|}{300} \\
\midrule
\endfirsthead
\caption[]{ANES (MIP) - Emotionality Classification Performance Metrics (GPT 4o) - (-,-,+) (continued)}\\
\toprule
Class & Precision & Recall & F1 & Precision & Recall & F1 & Precision & Recall & F1 & Precision & Recall & F1 \\
 & \multicolumn{3}{c|}{1} & \multicolumn{3}{c|}{10} & \multicolumn{3}{c|}{100} & \multicolumn{3}{c|}{300} \\
\midrule
\endhead
\midrule
\multicolumn{13}{r}{\textit{Continued on next page}}\\
\midrule
\endfoot
\bottomrule
\endlastfoot

0 & 0.942 & 0.916 & 0.929 & 0.948 & 0.862 & 0.903 & 0.938 & 0.880 & 0.908 & 0.936 & 0.898 & 0.916 \\
1 & 0.751 & 0.817 & 0.783 & 0.657 & 0.847 & 0.740 & 0.679 & 0.813 & 0.740 & 0.718 & 0.799 & 0.751 \\
\end{longtable}

\subsection*{ANES (MIP) - Emotionality Classification (-,+,-)}
\setlength{\tabcolsep}{3pt}
\begin{longtable}{|l|c|c|c|c|c|c|c|c|c|c|c|c|}
\caption{ANES (MIP) - Emotionality Classification Performance Metrics (GPT 4o) - (-,+,-)}\\
\toprule
Class & Precision & Recall & F1 & Precision & Recall & F1 & Precision & Recall & F1 & Precision & Recall & F1 \\
 & \multicolumn{3}{c|}{1} & \multicolumn{3}{c|}{10} & \multicolumn{3}{c|}{100} & \multicolumn{3}{c|}{300} \\
\midrule
\endfirsthead
\caption[]{ANES (MIP) - Emotionality Classification Performance Metrics (GPT 4o) - (-,+,-) (continued)}\\
\toprule
Class & Precision & Recall & F1 & Precision & Recall & F1 & Precision & Recall & F1 & Precision & Recall & F1 \\
 & \multicolumn{3}{c|}{1} & \multicolumn{3}{c|}{10} & \multicolumn{3}{c|}{100} & \multicolumn{3}{c|}{300} \\
\midrule
\endhead
\midrule
\multicolumn{13}{r}{\textit{Continued on next page}}\\
\midrule
\endfoot
\bottomrule
\endlastfoot

0 & 0.967 & 0.756 & 0.849 & 0.943 & 0.852 & 0.895 & 0.952 & 0.847 & 0.896 & 0.952 & 0.883 & 0.916 \\
1 & 0.539 & 0.918 & 0.679 & 0.637 & 0.834 & 0.722 & 0.637 & 0.861 & 0.732 & 0.695 & 0.856 & 0.767 \\
\end{longtable}

\subsection*{ANES (MIP) - Emotionality Classification (+,-,-)}
\setlength{\tabcolsep}{3pt}
\begin{longtable}{|l|c|c|c|c|c|c|c|c|c|c|c|c|}
\caption{ANES (MIP) - Emotionality Classification Performance Metrics (GPT 4o) - (+,-,-)}\\
\toprule
Class & Precision & Recall & F1 & Precision & Recall & F1 & Precision & Recall & F1 & Precision & Recall & F1 \\
 & \multicolumn{3}{c|}{1} & \multicolumn{3}{c|}{10} & \multicolumn{3}{c|}{100} & \multicolumn{3}{c|}{300} \\
\midrule
\endfirsthead
\caption[]{ANES (MIP) - Emotionality Classification Performance Metrics (GPT 4o) - (+,-,-) (continued)}\\
\toprule
Class & Precision & Recall & F1 & Precision & Recall & F1 & Precision & Recall & F1 & Precision & Recall & F1 \\
 & \multicolumn{3}{c|}{1} & \multicolumn{3}{c|}{10} & \multicolumn{3}{c|}{100} & \multicolumn{3}{c|}{300} \\
\midrule
\endhead
\midrule
\multicolumn{13}{r}{\textit{Continued on next page}}\\
\midrule
\endfoot
\bottomrule
\endlastfoot

0 & 0.953 & 0.888 & 0.919 & 0.957 & 0.815 & 0.880 & 0.987 & 0.727 & 0.837 & 0.983 & 0.724 & 0.834 \\
1 & 0.704 & 0.859 & 0.774 & 0.596 & 0.881 & 0.711 & 0.524 & 0.968 & 0.680 & 0.520 & 0.959 & 0.674 \\
\end{longtable}

\subsection*{ANES (MIP) - Emotionality Classification (-,+,+)}
\setlength{\tabcolsep}{3pt}
\begin{longtable}{|l|c|c|c|c|c|c|c|c|c|c|c|c|}
\caption{ANES (MIP) - Emotionality Classification Performance Metrics (GPT 4o) - (-,+,+)}\\
\toprule
Class & Precision & Recall & F1 & Precision & Recall & F1 & Precision & Recall & F1 & Precision & Recall & F1 \\
 & \multicolumn{3}{c|}{1} & \multicolumn{3}{c|}{10} & \multicolumn{3}{c|}{100} & \multicolumn{3}{c|}{300} \\
\midrule
\endfirsthead
\caption[]{ANES (MIP) - Emotionality Classification Performance Metrics (GPT 4o) - (-,+,+) (continued)}\\
\toprule
Class & Precision & Recall & F1 & Precision & Recall & F1 & Precision & Recall & F1 & Precision & Recall & F1 \\
 & \multicolumn{3}{c|}{1} & \multicolumn{3}{c|}{10} & \multicolumn{3}{c|}{100} & \multicolumn{3}{c|}{300} \\
\midrule
\endhead
\midrule
\multicolumn{13}{r}{\textit{Continued on next page}}\\
\midrule
\endfoot
\bottomrule
\endlastfoot

0 & 0.896 & 0.961 & 0.927 & 0.896 & 0.914 & 0.905 & 0.891 & 0.952 & 0.920 & 0.899 & 0.941 & 0.920 \\
1 & 0.837 & 0.639 & 0.725 & 0.704 & 0.658 & 0.680 & 0.801 & 0.623 & 0.701 & 0.777 & 0.658 & 0.712 \\
\end{longtable}

\subsection*{ANES (MIP) - Emotionality Classification (+,-,+)}
\setlength{\tabcolsep}{3pt}
\begin{longtable}{|l|c|c|c|c|c|c|c|c|c|c|c|c|}
\caption{ANES (MIP) - Emotionality Classification Performance Metrics (GPT 4o) - (+,-,+)}\\
\toprule
Class & Precision & Recall & F1 & Precision & Recall & F1 & Precision & Recall & F1 & Precision & Recall & F1 \\
 & \multicolumn{3}{c|}{1} & \multicolumn{3}{c|}{10} & \multicolumn{3}{c|}{100} & \multicolumn{3}{c|}{300} \\
\midrule
\endfirsthead
\caption[]{ANES (MIP) - Emotionality Classification Performance Metrics (GPT 4o) - (+,-,+) (continued)}\\
\toprule
Class & Precision & Recall & F1 & Precision & Recall & F1 & Precision & Recall & F1 & Precision & Recall & F1 \\
 & \multicolumn{3}{c|}{1} & \multicolumn{3}{c|}{10} & \multicolumn{3}{c|}{100} & \multicolumn{3}{c|}{300} \\
\midrule
\endhead
\midrule
\multicolumn{13}{r}{\textit{Continued on next page}}\\
\midrule
\endfoot
\bottomrule
\endlastfoot

0 & 0.938 & 0.892 & 0.915 & 0.945 & 0.823 & 0.880 & 0.946 & 0.843 & 0.892 & 0.942 & 0.875 & 0.907 \\
1 & 0.700 & 0.810 & 0.751 & 0.597 & 0.844 & 0.699 & 0.626 & 0.846 & 0.719 & 0.672 & 0.827 & 0.742 \\
\end{longtable}

\subsection*{ANES (MIP) - Emotionality Classification (+,+,-)}
\setlength{\tabcolsep}{3pt}
\begin{longtable}{|l|c|c|c|c|c|c|c|c|c|c|c|c|}
\caption{ANES (MIP) - Emotionality Classification Performance Metrics (GPT 4o) - (+,+,-)}\\
\toprule
Class & Precision & Recall & F1 & Precision & Recall & F1 & Precision & Recall & F1 & Precision & Recall & F1 \\
 & \multicolumn{3}{c|}{1} & \multicolumn{3}{c|}{10} & \multicolumn{3}{c|}{100} & \multicolumn{3}{c|}{300} \\
\midrule
\endfirsthead
\caption[]{ANES (MIP) - Emotionality Classification Performance Metrics (GPT 4o) - (+,+,-) (continued)}\\
\toprule
Class & Precision & Recall & F1 & Precision & Recall & F1 & Precision & Recall & F1 & Precision & Recall & F1 \\
 & \multicolumn{3}{c|}{1} & \multicolumn{3}{c|}{10} & \multicolumn{3}{c|}{100} & \multicolumn{3}{c|}{300} \\
\midrule
\endhead
\midrule
\multicolumn{13}{r}{\textit{Continued on next page}}\\
\midrule
\endfoot
\bottomrule
\endlastfoot

0 & 0.946 & 0.884 & 0.914 & 0.935 & 0.876 & 0.905 & 0.914 & 0.890 & 0.902 & 0.946 & 0.852 & 0.896 \\
1 & 0.691 & 0.837 & 0.757 & 0.668 & 0.805 & 0.730 & 0.673 & 0.730 & 0.700 & 0.639 & 0.843 & 0.727 \\
\end{longtable}

\subsection*{ANES (MIP) - Emotionality Classification (+,+,+)}
\setlength{\tabcolsep}{3pt}
\begin{longtable}{|l|c|c|c|c|c|c|c|c|c|c|c|c|}
\caption{ANES (MIP) - Emotionality Classification Performance Metrics (GPT 4o) - (+,+,+)}\\
\toprule
Class & Precision & Recall & F1 & Precision & Recall & F1 & Precision & Recall & F1 & Precision & Recall & F1 \\
 & \multicolumn{3}{c|}{1} & \multicolumn{3}{c|}{10} & \multicolumn{3}{c|}{100} & \multicolumn{3}{c|}{300} \\
\midrule
\endfirsthead
\caption[]{ANES (MIP) - Emotionality Classification Performance Metrics (GPT 4o) - (+,+,+) (continued)}\\
\toprule
Class & Precision & Recall & F1 & Precision & Recall & F1 & Precision & Recall & F1 & Precision & Recall & F1 \\
 & \multicolumn{3}{c|}{1} & \multicolumn{3}{c|}{10} & \multicolumn{3}{c|}{100} & \multicolumn{3}{c|}{300} \\
\midrule
\endhead
\midrule
\multicolumn{13}{r}{\textit{Continued on next page}}\\
\midrule
\endfoot
\bottomrule
\endlastfoot

0 & 0.905 & 0.951 & 0.927 & 0.922 & 0.897 & 0.909 & 0.901 & 0.908 & 0.904 & 0.927 & 0.910 & 0.919 \\
1 & 0.811 & 0.677 & 0.738 & 0.695 & 0.754 & 0.723 & 0.695 & 0.678 & 0.686 & 0.728 & 0.769 & 0.748 \\
\end{longtable}
\end{document}